\documentclass[11pt]{article}

\usepackage{graphicx}
\usepackage{amsmath, amssymb}
\usepackage{enumitem}
\usepackage{cite}
\usepackage{bbm}
\usepackage{authblk}

\usepackage{booktabs}
\usepackage{multirow}
\usepackage{subcaption}
\usepackage{pdflscape}
\usepackage{xurl}
\usepackage[pagebackref=true,breaklinks=true,colorlinks,bookmarks=false]{hyperref}

\title{\textbf{Deep learning from routine histology improves risk stratification for biochemical recurrence in prostate cancer}}

\author[1]{Clément Grisi\footnote{corresponding author: clement.grisi@radboudumc.nl}}
\author[1]{Khrystyna Faryna}
\author[1]{Nefise Uysal}
\author[1,2]{Vittorio Agosti}
\author[2,3]{Enrico Munari}
\author[4]{Solène-Florence Kammerer-Jacquet}
\author[5]{Paulo Guilherme de Oliveira Salles}
\author[6]{Yuri Tolkach}
\author[6]{Reinhard Büttner}
\author[7]{Sofiya Semko}
\author[7]{Maksym Pikul}
\author[7]{Axel Heidenreich}
\author[1]{Jeroen van der Laak}
\author[1]{Geert Litjens}
\affil[1]{Department of Pathology, Radboud University Medical Center, Nijmegen, The Netherlands}
\affil[2]{Department of Molecular and Translational Medicine, University of Brescia, Brescia, Italy}
\affil[3]{Surgical Pathology Unit, Verona University Hospital Trust, Verona, Italy}
\affil[4]{Department of Pathology, CHU de Rennes, Rennes, France}
\affil[5]{Anatomical Pathology Service, Instituto Mário Penna, Belo Horizonte, Brazil}
\affil[6]{Institute of Pathology, University Hospital Cologne, Cologne, Germany}
\affil[7]{Clinic of Urology, University Hospital Cologne, Cologne, Germany}

\date{}

\begin{document}
	\maketitle
	
	\begin{abstract}
		Accurate prediction of biochemical recurrence (BCR) after radical prostatectomy is critical for guiding adjuvant treatment and surveillance decisions in prostate cancer. However, existing clinicopathological risk models reduce complex morphology to relatively coarse descriptors, leaving substantial prognostic information embedded in routine histopathology underexplored. We present a deep learning-based biomarker that predicts continuous, patient-specific risk of BCR directly from H\&E-stained whole-slide prostatectomy specimens. Trained end-to-end on time-to-event outcomes and evaluated across four independent international cohorts, our model demonstrates robust generalization across institutions and patient populations. When integrated with the CAPRA-S clinical risk score, our deep learning risk score consistently improved discrimination for BCR, increasing concordance indices from $0.725$--$0.772$ to $0.749$--$0.788$ across cohorts. To support clinical interpretability, outcome-grounded analyses revealed subtle histomorphological patterns associated with recurrence risk that are not captured by conventional clinicopathological risk scores. This multicohort study demonstrates that deep learning applied to routine prostate histopathology can deliver reproducible and clinically generalizable biomarkers that augment postoperative risk stratification, with potential to support personalized management of prostate cancer in real-world clinical settings.
	\end{abstract}

	\section{Introduction}
	
	Prostate cancer is the second most commonly diagnosed cancer worldwide and the fifth leading cause of cancer-related death, with more than 1.5 million men affected annually \cite{Bray2024}. Many patients undergo radical prostatectomy, either as primary treatment or following a period of active surveillance. The success of surgery is routinely monitored through the concentration of prostate-specific antigen (PSA) in the blood, which is expected to drop to undetectable levels ($<0.1$ ng/mL) within 4 to 6 weeks after surgery. However, in approximately $30\%$ of the patients, PSA levels subsequently rise, indicating malignant cell regrowth and portending an increased risk of metastasis and disease-specific mortality. Biochemical recurrence (BCR), historically defined as PSA exceeding $0.2$ ng/mL on two or more consecutive occasions after reaching nadir, is therefore an early indicator of treatment failure preceding overt clinical progression. Accurate identification of patients at risk for BCR is central to postoperative decision-making, informing the use of adjuvant radiotherapy, advanced imaging, genomic testing, and the intensity of long-term surveillance.\\
	\\
	To support these decisions, several statistical risk models combining clinical and pathological variables have been developed \cite{Cooperberg2005,Sandeman2020}. Among these, CAPRA-S is widely adopted and has demonstrated strong prognostic performance across multiple cohorts \cite{Cooperberg2011}. Nevertheless, such models rely on manual histopathological assessment and structured reporting, which limits reproducibility and constrains the granularity at which morphology is represented. Central to most risk models is Gleason grading, which not only suffers from substantial inter-observer variability \cite{Ozkan2016,VanLeenders2020} but also reduces spatially heterogeneous tumor morphology into coarse categorical descriptors. As a result, within-grade architectural patterns and subtle histological features that may carry independent prognostic information are currently overlooked, leaving a substantial fraction of the information embedded in routine histopathology unexplored.\\
	\\
	Deep learning has the potential to unlock the true prognostic value of morphological assessment in cancer by capturing fine-grained, spatially distributed histological patterns that are not routinely assessed or quantified in clinical practice. Early approaches based on handcrafted features \cite{Yamamoto2019} or simplistic models \cite{Leo2021} did not directly optimize on patient outcomes, limiting their ability to discover clinically meaningful prognostic signals. More recent end-to-end models \cite{Pinckaers2022,dietrich21a,Farrokh2024} have shown improved performance but were largely restricted to tissue microarrays or biopsy cores, which fail to capture the full spatial and morphological heterogeneity present in prostatectomy specimens. In contrast, whole-slide images (WSIs) provide a more comprehensive view of tumor biology, but their gigapixel size poses unique computational challenges. Weakly supervised learning frameworks, particularly multiple instance learning (MIL), have emerged as effective solutions, enabling outcome-driven training on WSIs without pixel-level annotations \cite{Campanella2019,Hou2016,Ilse2018,Courtiol2019,Lu2021}. Incorporating spatial context into these models has further improved performance in grading and prognostic tasks \cite{lerousseau21a,shao2021,chen2022,Grisi2025}.\\
	\\
	In parallel, advances in self-supervised learning have ushered in the era of pathology foundation models. Pretrained on large, diverse collections of unlabeled histology images, these encoders learn rich, domain-specific representations that transfer effectively to downstream tasks \cite{Wang2022,Chen2024}. While these models promise improved robustness and generalization, systematic evaluations in clinically relevant settings remain scarce.\\
	\\
	In this study, we present a deep learning approach that predicts risk of BCR directly from whole prostatectomy histology using a continuous time-to-event endpoint. We evaluate three large pathology foundation models alongside a smaller, prostatectomy-specific encoder within a context-aware MIL framework explicitly trained on postoperative recurrence outcomes. Trained on the largest dataset yet assembled for this task, the best models demonstrate robust generalization across four independent international cohorts. Our results show that image-derived risk scores provide prognostic information complementary to CAPRA-S, demonstrating that computational pathology can deliver robust, reproducible biomarkers that augment postoperative risk stratification in prostate cancer.	
	
	\section{Results}
	
	\subsection{Data overview and experimental setup}
	
\begin{figure}[!h]
		\centering
		\includegraphics[width=\textwidth]{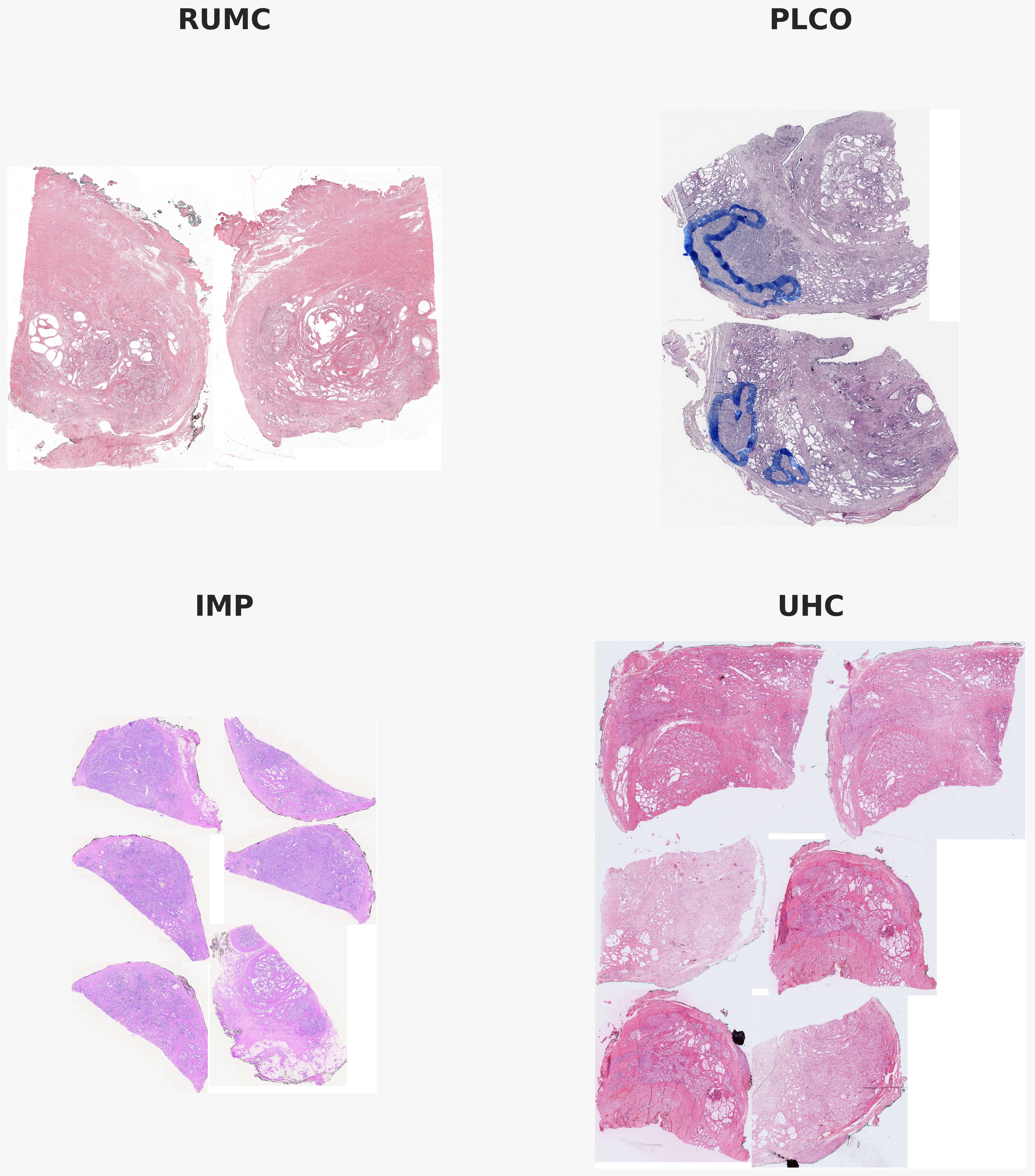}
		\caption{\textbf{Representative whole-slide histopathology from external test cohorts.}	Low-resolution thumbnails of H\&E-stained prostatectomies from the four external test cohorts used in this study. The examples illustrate cohort-specific differences in tissue appearance and color profiles arising from variations in staining protocols, slide preparation, and slide acquisition. All slides were processed using the same preprocessing pipeline prior to model inference.}
		\label{fig:wsi-thumb}
	\end{figure}

\begin{figure}[!htbp]
		\centering
		\includegraphics[width=\textwidth]{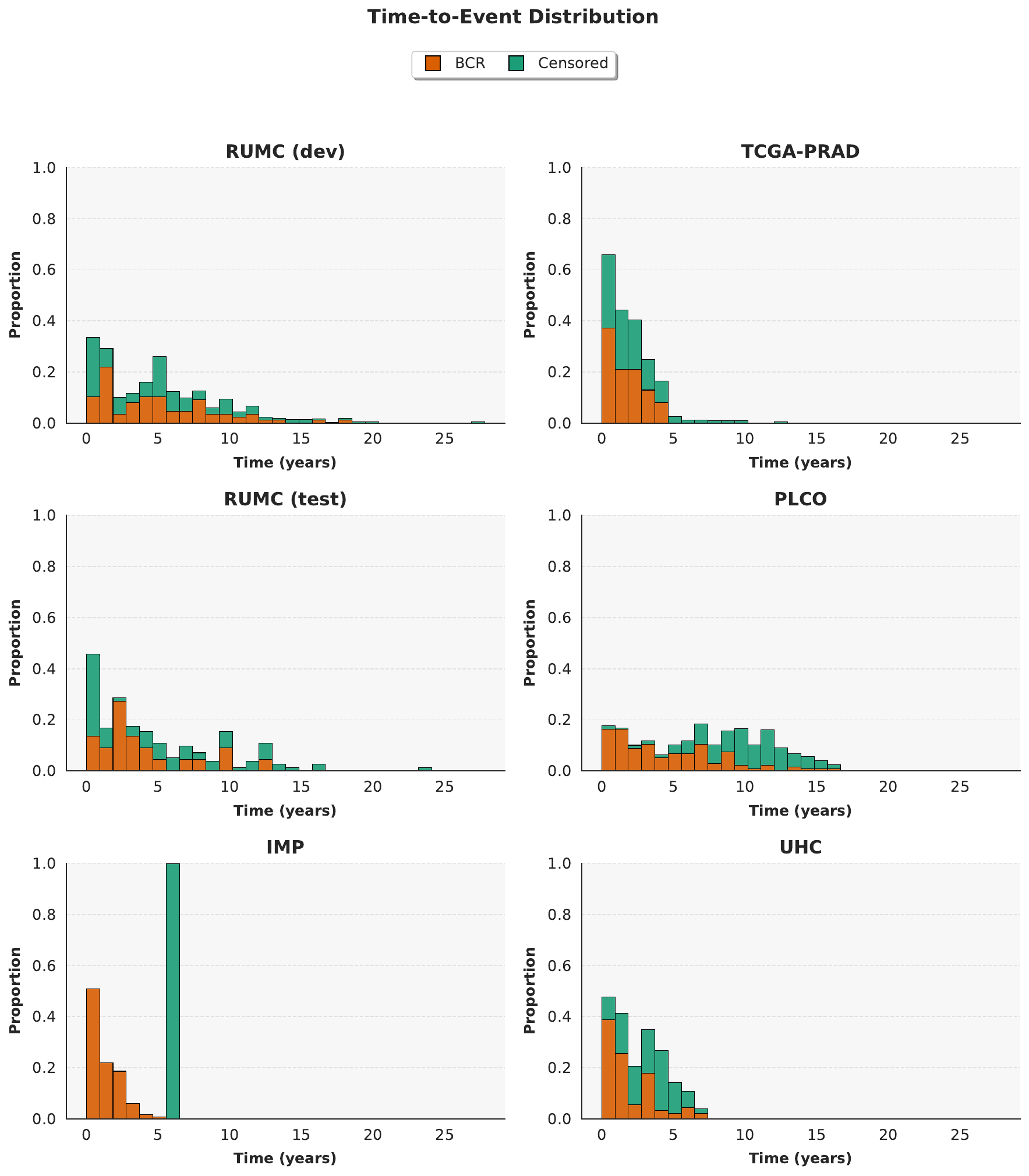}
		\caption{\textbf{Time-to-event distributions across cohorts.} Stacked histograms show times from radical prostatectomy to biochemical recurrence or last follow-up in each cohort used in this study. The horizontal axis represents time in years, and the vertical axis shows the proportion of patients within each cohort, subdivided into BCR (orange) and censored (green) cases. Consistent bin edges are used to facilitate comparison across cohorts. Cohort-specific differences in follow-up times reflect variations in recruitment periods, surveillance protocols, and clinical practices.}
		\label{fig:label}
	\end{figure}

	We used five independent cohorts encompassing both historical and contemporary clinical practice, comprising a total of $2485$ patients and $7262$ whole-slide images and spanning surgeries from 1992 to 2020 across Europe, North America, and South America. This multicohort design covers a wide range of patient populations, follow-up protocols, and slide acquisition systems, enabling a controlled assessment of how \emph{training data heterogeneity} affects model generalization under real-world conditions. The main characteristics of each dataset are summarized in Table \ref{tab:dataset-summary}. Figure \ref{fig:wsi-thumb} illustrates representative whole-slide image thumbnails from each external test cohort, highlighting differences in tissue appearance and color profiles arising from cohort-specific staining protocols and slide acquisition systems. The distribution of International Society of Urological Pathology (ISUP) grades for each cohort is shown in Figure \ref{fig:isup}, while Figure \ref{fig:label} shows the distribution of biochemical recurrence and censoring times. 
	
	\begin{table}[ht]
		\centering
		\caption{\textbf{Overview of datasets}. Summary of the main characteristics of each cohort used in this study, including collection period, number of patients, number of associated digital pathology slides, and scanner system used for digitization. \texttt{TCGA-PRAD} comprises cases from multiple contributing institutions, predominantly from the United States.}
		\label{tab:dataset-summary}
		\resizebox{\textwidth}{!}{\begin{tabular}{cccccc}
				\toprule
				\textbf{Cohort} & \textbf{Location} & \textbf{Years of surgery} & \textbf{\# Patients} & \textbf{\# Slides} & \textbf{Scanner} \\
				\midrule
				\texttt{RUMC} & Nijmegen, The Netherlands & 1992--2012 & 608 & 1723 & 3DHistech P1000 \\[0.3em]
				\texttt{PLCO} & United States & 1993--2001 & 723 & 1971 & Leica \\[0.3em]
				\texttt{IMP} & Belo Horizonte, Brazil & 2016 & 421 & 2091 & Motic Easy Scan Infinity 60N \\[0.3em]
				\texttt{UHC} & Cologne, Germany & 2015--2020 & 330 & 1028 & Hamamatsu NanoZoomer S360 \\[0.3em]
				\texttt{TCGA-PRAD} & \textit{Multiple locations} & 2000--2013 & 403 & 449 & \textit{Multiple scanners} \\
				\bottomrule
		\end{tabular}}
	\end{table}
	
	\paragraph{Model development.} Model training was based on $508$ patients from the \texttt{RUMC} cohort, publicly released through the \texttt{LEOPARD} (LEarning biOchemical Prostate cAncer Recurrence from histopathology sliDes) challenge. To isolate the effect of \emph{training-set heterogeneity} while keeping the downstream architecture fixed, we defined two model development regimes. The first, \texttt{RUMC-only}, used the $508$ \texttt{RUMC} patients. The second, \texttt{RUMC+TCGA}, enriched this cohort with $298$ \texttt{TCGA-PRAD} patients, introducing substantial technical variation (e.g., staining and scanners) and a distinct clinical mix characteristic of a multi-institution resource, for a total of $806$ patients. Both configurations were split into five cross-validation folds using a stratified procedure to preserve the distribution of recurrence status, Gleason grades, and year of surgery (Section \ref{sec:preprocessing}).
	
	\paragraph{External validation.} Model generalization was assessed on four independent cohorts: $100$ patients from \texttt{RUMC}, $723$ from \texttt{PLCO}, $421$ from \texttt{IMP}, and $330$ from \texttt{UHC}. While the \texttt{RUMC} test set is closely related to the training data, the remaining cohorts differ substantially in both technical and clinical characteristics, providing a stringent evaluation of out-of-domain generalization. Importantly, this evaluation spans both historical and contemporary clinical practice. The \texttt{RUMC}, \texttt{PLCO}, and \texttt{IMP} cohorts adopt the historical definition of biochemical recurrence based on two consecutive PSA measurements $\geq 0.2$ ng/mL, whereas \texttt{UHC} reflects contemporary practice, defining recurrence as a single postoperative PSA measurement $\geq 0.1$ ng/mL. This endpoint shift serves as a clinically meaningful test of deployability, assessing whether models trained on historical criteria remain predictive under a contemporary recurrence definition (Section \ref{sec:materials}).
	
	\paragraph{Data preprocessing.} For each patient, diagnostic slides were combined into a single patient-level image and tiled into non-overlapping regions at $0.50$ microns per pixel (mpp), enabling unified inference across multiple slides while preserving spatially distributed signals present throughout the prostatectomy specimen. This resolution was selected to retain fine-grained morphological detail relevant for risk prediction while remaining compatible with the operating conditions of the encoders evaluated in this study. To avoid masking potentially prognostic cues at the interface between malignant and non-malignant tissue, all available tissue, including benign regions, was retained during tiling, allowing the model to leverage both tumor morphology and contextual non-tumor features potentially associated with recurrence. To further account for scanner variability, a $5$\% tolerance was applied. When no native spacing fell within this range, regions were extracted at the nearest higher resolution and resampled to approximate $0.50$ mpp (Methods, Section \ref{sec:preprocessing}).
	
	\subsection{Pretraining diversity drives generalization when models are trained on data from a single source}
	\label{sec:pretraining-diversity}
	
\begin{figure}[!h]
		\centering
		\includegraphics[width=\textwidth]{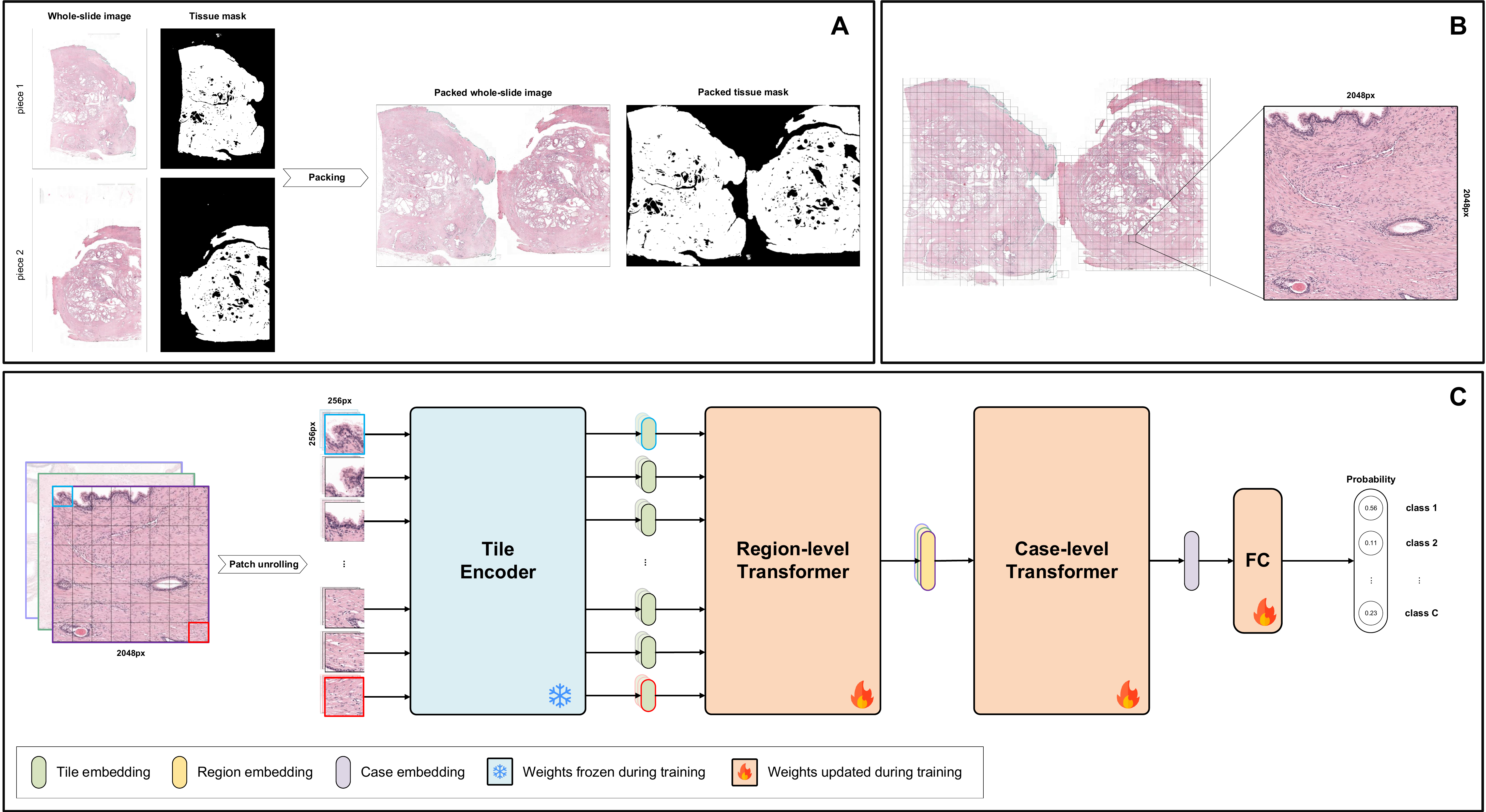}
		\caption{\textbf{Integration of multiple histopathology slides into a unified patient-level prediction framework.} \textbf{(A)} For patients with multiple slides, segmented tissue regions are cropped using the corresponding tissue masks and stitched into a single larger \textit{packed} slide. This packing process minimizes empty background space while preserving the relative arrangement of tissue, enabling patient-level analysis in a unified coordinate frame. \textbf{(B)} The packed slide is tiled into non-overlapping square regions of $2048$ pixels at $0.50$ mpp. Regions with less than $1$\% tissue coverage, estimated from the segmentation mask, are excluded. \textbf{(C)} Each retained region is then unrolled into non-overlapping $256$ pixel tiles, which are processed by a frozen tile encoder to generate tile-level embeddings. For each region, the sequence of tile embeddings is aggregated into a region-level embedding by a first Transformer. The sequence of region-level embeddings for the entire case is then aggregated by a second Transformer into a single patient-level representation, which is projected onto the target classes via a fully connected (FC) layer. The two Transformers and the FC layer are jointly trained for biochemical recurrence risk prediction. Together, these steps enable efficient processing of large whole-slide images, integrating spatially distributed information from multiple slides into a unified patient-level prediction.}
		\label{fig:pipeline}
	\end{figure}

	To assess how pretraining diversity influences generalization when downstream training data originate from a single institution, we evaluated multiple vision encoders within a common weakly supervised learning framework for predicting biochemical recurrence from whole-slide images. All models were implemented using a context-aware multiple instance learning pipeline based on the Hierarchical Image Pyramid Transformer (HIPT) \cite{chen2022}, a three-stage architecture that aggregates information at progressively coarser spatial resolutions (Figure \ref{fig:pipeline}\textbf{C}; Methods, Section \ref{sec:hipt}). Within this fixed downstream architecture, we compared four encoders: \texttt{Prost40M}, a prostatectomy-specific encoder (Methods, Section \ref{sec:dino-vit-s}), and three large pathology foundation models -- \texttt{UNI} \cite{Chen2024}, \texttt{Virchow2} \cite{zimmermann2024}, and \texttt{H-optimus-0} \cite{hoptimus0}. Key encoder characteristics are summarized in Table \ref{tab:encoder-summary}.\\
	\\
	Model training used discrete-time survival modeling that enables training with a batch size of one, which is well suited for gigapixel whole-slide inputs. The continuous time scale was partitioned into four non-overlapping intervals defined by quartiles of observed recurrence times among uncensored patients, and models were optimized using a censoring-aware discrete-time log-likelihood (Methods, Section \ref{sec:discrete-time-modelling}). To isolate the effect of encoder pretraining under homogeneous downstream data, all models were first trained and evaluated using stratified 5-fold cross-validation on the $508$ \texttt{RUMC} patients (\texttt{RUMC-only} splits). Performance was reported using the mean concordance index across folds, and patient-level ensemble risk scores were obtained by averaging predictions from the five fold-specific models for subsequent statistical testing.
	
	\begin{table*}[ht]
		\centering
		\caption{\textbf{Overview of vision encoders}. Summary of the four vision encoders evaluated in this study, including model architecture, number of parameters, pretraining corpus size and source, and pretraining method.}
		\label{tab:encoder-summary}
		\resizebox{\textwidth}{!}{\begin{tabular}{ccccc}
				\toprule
				\textbf{Encoder} & \textbf{Architecture} & \textbf{Parameters} & \textbf{Pretraining corpus} & \textbf{Pretraining method} \\
				\midrule
				\texttt{Prost40M} & ViT-Small & 22M & 40M prostatectomy patches $\sim$2k WSIs & DINO \\[0.3em]
				\texttt{UNI} & ViT-Large & 307M & 100M patches from $>$100k WSIs & DINOv2 \\[0.3em]
				\texttt{Virchow2} & ViT-Huge & 632M & 1.9B patches from 3.1M WSIs & DINOv2 \\[0.3em]
				\texttt{H-optimus-0} & ViT-giant & 1.1B & 273M patches from 500k WSIs & DINOv2 \\
				\bottomrule
		\end{tabular}}
	\end{table*}
	
	\vspace{4mm}
	
\begin{figure}[!htbp]
		\centering
		\includegraphics[width=\linewidth]{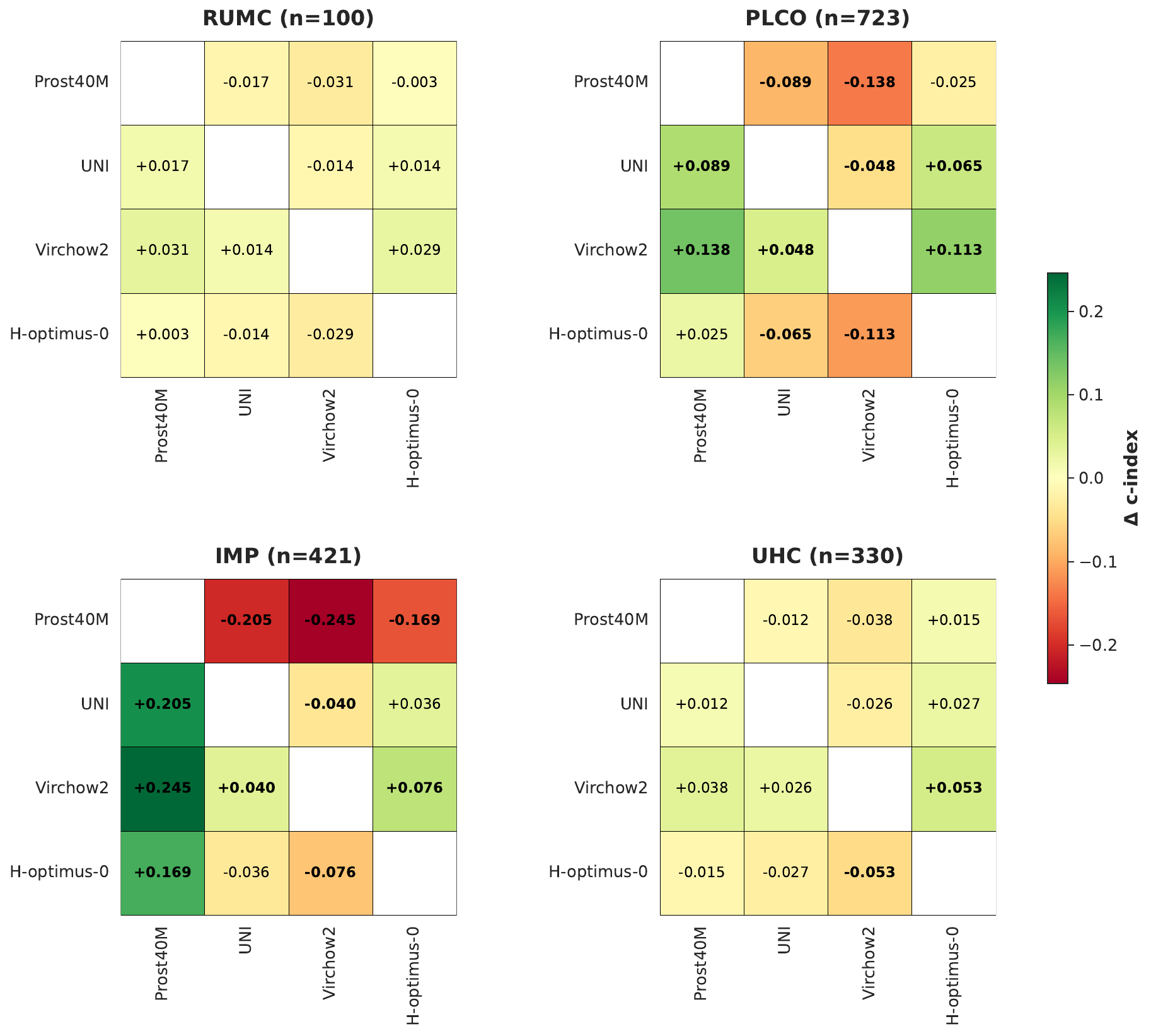}
		\caption{\textbf{Pairwise statistical comparisons of encoder performance across test cohorts.} Heatmaps depict differences in concordance index between models trained on the \texttt{RUMC-only} splits and evaluated on four independent test cohorts (\texttt{RUMC}, \texttt{PLCO}, \texttt{IMP}, \texttt{UHC}). Each cell indicates the difference in performance between the encoder in the row and the encoder in the column. Positive values (green) reflect higher performance for the encoder in the row relative to the encoder in the column, while negative values (red) indicate the opposite. Significant differences are highlighted in bold.}
		\label{fig:heatmaps-rumc}
	\end{figure}

	\noindent
	Internal cross-validation performance was uniformly high, with mean c-indices ranging from 0.684 (\texttt{H-optimus-0}) to 0.727 (\texttt{UNI}), indicating that all encoders achieved strong performance when evaluated on data drawn from the same distribution as the training set (Table \ref{table:rumc-only}). Evaluation on held-out test cohorts, however, revealed marked differences in generalization. \texttt{Prost40M} performed reasonably well on in-domain \texttt{RUMC} (0.646) but showed substantial performance degradation on external cohorts (0.573 on \texttt{PLCO}, 0.496 on \texttt{IMP}, 0.626 on \texttt{UHC}). Foundation models exhibited stronger and more consistent generalization to out-of-domain data, with \texttt{Virchow2} emerging as the most robust encoder across external cohorts (0.701 on \texttt{PLCO}, 0.663 on \texttt{IMP}, and 0.671 on \texttt{UHC}). Pairwise statistical testing confirmed these trends (Figure \ref{fig:heatmaps-rumc}, Appendix \ref{app:pairwise-comparisons} Table \ref{table:pairwise-rumc}). While no significant differences were observed on the internal \texttt{RUMC} test set (all $q > 0.15$), clear performance gaps emerged externally: \texttt{Virchow2} significantly outperformed all other encoders in both \texttt{PLCO} and \texttt{IMP}, and showed a significant advantage over \texttt{H-optimus-0} in \texttt{UHC} ($q < 0.01$). \\
	\\
	Together, these results indicate that when downstream training data are homogeneous, robustness is strongly influenced by the heterogeneity encountered during pretraining. In this setting, pathology foundation models are preferable to narrow, domain-specific encoders, as their representations are less biased toward a single source domain. Among them, \texttt{Virchow2} demonstrated the strongest generalization, consistent with its exceptionally large and diverse pretraining corpus, which provides stronger inductive biases than the more limited pretraining of \texttt{UNI} or \texttt{H-optimus-0}.
	
	\begin{table}[htbp]
		\centering
		
		\begin{minipage}{\textwidth}
			\caption{\textbf{Cross-validation performance of deep learning models.} Five-fold cross-validation c-index (mean ± std) of models trained on different data splits and evaluated on held-out test cohorts. Best values for each cohort are shown in bold.}
			\vspace{0.5em}
			
			\begin{subtable}[t]{\textwidth}
				\centering
				\caption{Training on \texttt{RUMC-only} splits}
				\resizebox{\textwidth}{!}{%
					\begin{tabular}{lccccc}
						\toprule
						\textbf{Encoder} & \textbf{Tuning} & \textbf{RUMC} & \textbf{PLCO} & \textbf{IMP} & \textbf{UHC} \\
						\midrule
						\texttt{Prost40M} & 0.712 ± 0.069 & 0.646 ± 0.062 & 0.573 ± 0.035 & 0.496 ± 0.056 & 0.626 ± 0.043 \\
						\texttt{UNI}          & \textbf{0.727 ± 0.040} & 0.676 ± 0.030 & 0.679 ± 0.009 & 0.605 ± 0.088 & 0.629 ± 0.048 \\
						\texttt{Virchow2}     & 0.694 ± 0.066 & \textbf{0.692 ± 0.040} & \textbf{0.701 ± 0.030} & \textbf{0.663 ± 0.058} & \textbf{0.671 ± 0.040} \\
						\texttt{H-optimus-0}  & 0.684 ± 0.024 & 0.674 ± 0.025 & 0.641 ± 0.045 & 0.619 ± 0.061 & 0.625 ± 0.032 \\
						\bottomrule
					\end{tabular}%
					\label{table:rumc-only}
				}
			\end{subtable}
			
			\vspace{1em}
			
			\begin{subtable}[t]{\textwidth}
				\centering
				\caption{Training on \texttt{RUMC+TCGA} splits}
				\resizebox{\textwidth}{!}{%
					\begin{tabular}{lccccc}
						\toprule
						\textbf{Encoder} & \textbf{Tuning} & \textbf{RUMC} & \textbf{PLCO} & \textbf{IMP} & \textbf{UHC} \\
						\midrule
						\texttt{Prost40M} & 0.723 ± 0.021 & \textbf{0.717 ± 0.019} & \textbf{0.685 ± 0.031} & \textbf{0.663 ± 0.057} & 0.682 ± 0.023 \\
						\texttt{UNI}          & 0.732 ± 0.025 & 0.663 ± 0.024 & 0.647 ± 0.072 & 0.583 ± 0.086 & 0.674 ± 0.032 \\
						\texttt{Virchow2}     & \textbf{0.749 ± 0.025} & 0.705 ± 0.073 & 0.655 ± 0.039 & 0.621 ± 0.070 & \textbf{0.713 ± 0.024} \\
						\texttt{H-optimus-0}  & 0.717 ± 0.025 & 0.647 ± 0.030 & 0.646 ± 0.037 & 0.576 ± 0.071 & 0.663 ± 0.020\\
						\bottomrule
					\end{tabular}%
					\label{table:rumc+tcga}
				}
			\end{subtable}
		\end{minipage}
	\end{table}
	
	\subsection{Conditional benefits of training data heterogeneity}
	
\begin{figure}[!htbp]
		\centering
		\includegraphics[width=\linewidth]{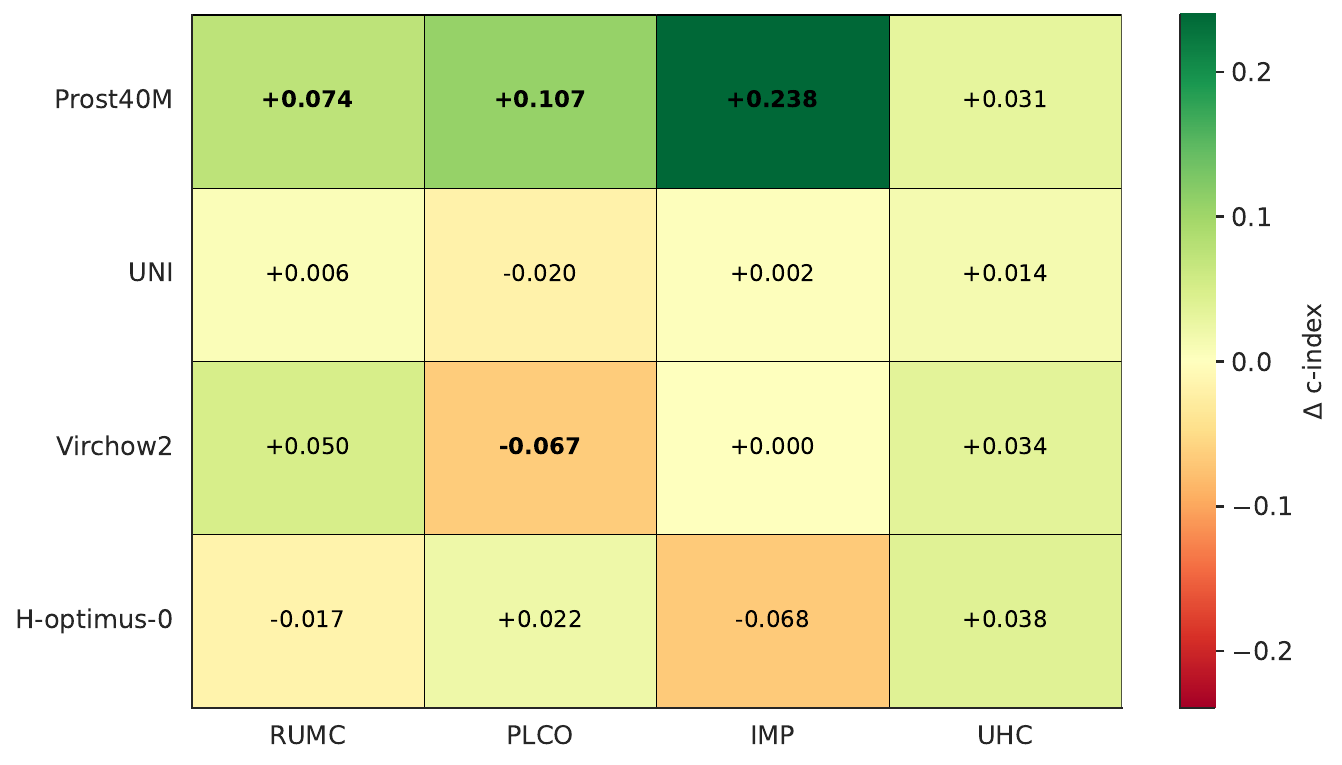}
		\caption{\textbf{Effect of training data enrichment on model performance across test cohorts.} The heatmap depicts the change in concordance index ($\Delta$) for each model when trained on the combined \texttt{RUMC+TCGA} dataset compared to training on \texttt{RUMC-only}. Rows correspond to encoders and columns to test cohorts. Positive values (green) indicate improved performance with training data enrichment, while negative values (red) indicate decreased performance. Significant differences are highlighted in bold.}
		\label{fig:heatmap-dataset}
	\end{figure}

\begin{figure}[!htbp]
		\centering
		\includegraphics[width=\linewidth]{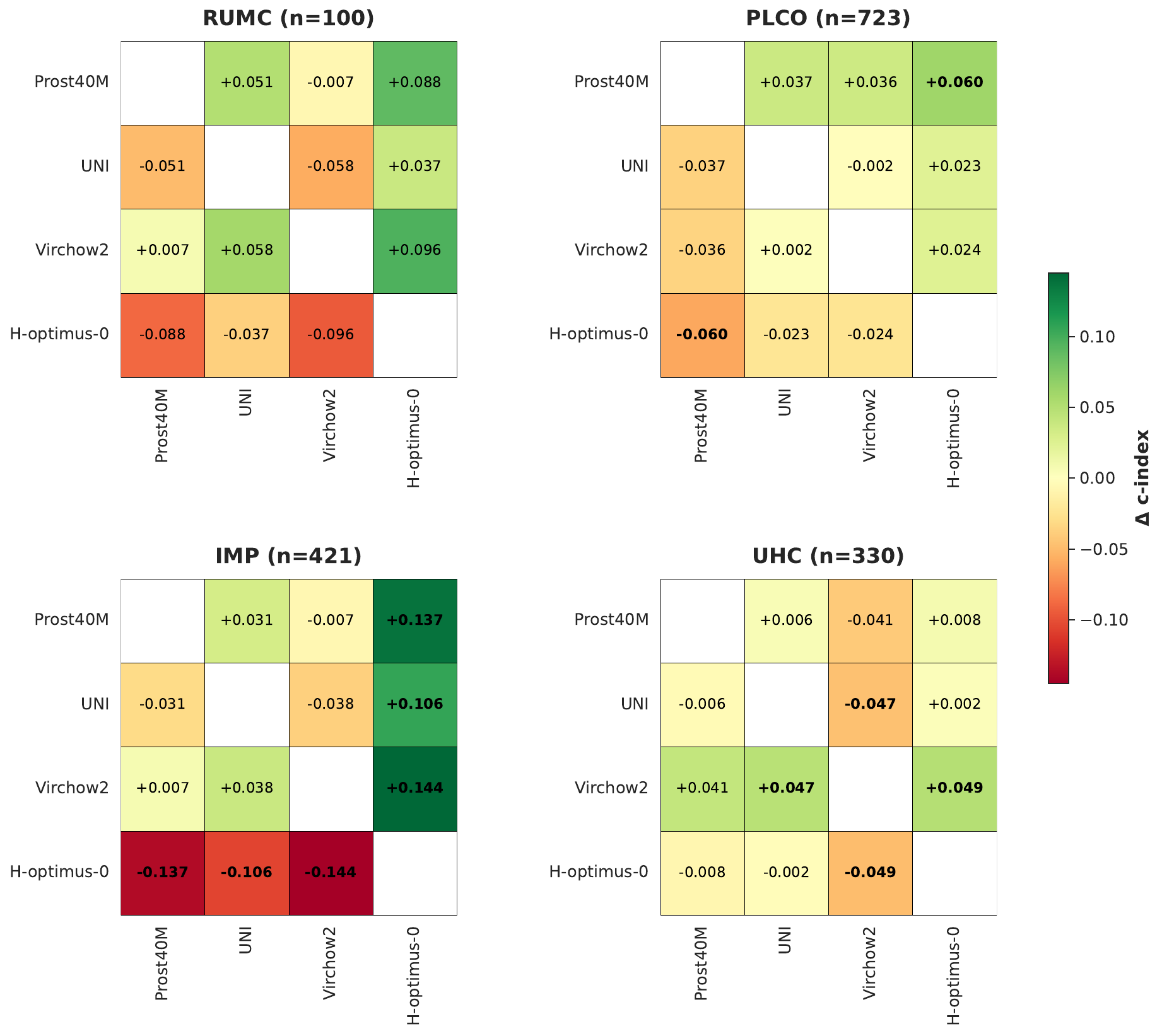}
		\caption{\textbf{Pairwise statistical comparisons of encoder performance across test cohorts.} Heatmaps depict differences in concordance index between models trained on the \texttt{RUMC+TCGA} splits and evaluated on four independent test cohorts (\texttt{RUMC}, \texttt{PLCO}, \texttt{IMP}, \texttt{UHC}). Each cell indicates the difference in performance between the encoder in the row and the encoder in the column. Positive values (green) reflect higher performance for the encoder in the row relative to the encoder in the column, while negative values (red) indicate the opposite. Significant differences are highlighted in bold.}
		\label{fig:heatmaps-rumc+tcga}
	\end{figure}

	Pretraining diversity proved critical for generalization when downstream training data were homogeneous. We next asked whether increasing \emph{training data heterogeneity} yields similar benefits, particularly for encoders with limited pretraining diversity. To this end, we enriched the $508$ \texttt{RUMC} cases with $298$ cases from \texttt{TCGA-PRAD}, introducing substantial technical heterogeneity (staining, scanners) as well as a distinct clinical mix characteristic of a multi-institution cohort. Models were evaluated using stratified 5-fold cross-validation on the enriched \texttt{RUMC+TCGA} splits, and results were confirmed through robust statistical testing (Methods, Section \ref{sec:statistics}).\\
	\\
	The effect of training data enrichment differed substantially between encoders (Figure \ref{fig:heatmap-dataset}, Appendix \ref{app:dataset-enrichement} Table \ref{table:dataset-enrichement-statistics}). The prostatectomy-specific encoder \texttt{Prost40M} exhibited large and statistically significant performance gains on three of the four external cohorts, suggesting that its narrow pretraining left it vulnerable to exploiting spurious, center-specific correlations when trained on \texttt{RUMC-only}. Adding \texttt{TCGA-PRAD} introduced sufficient technical and clinical variation to steer the model toward more generalizable prognostic features. In contrast, pathology foundation models derived little benefit from enrichment, which we interpret as a robustness effect: their broad and heterogeneous pretraining already yields representations that generalize reasonably well even when downstream training data are comparatively homogeneous. Accordingly, \texttt{Virchow2} remained strong overall but showed a decline on \texttt{PLCO} ($\Delta = -0.07$, $q < 0.01$), while \texttt{UNI} and \texttt{H-optimus-0} showed no measurable effect (all $q > 0.1$). Pairwise model comparisons corroborated these findings (Figure \ref{fig:heatmaps-rumc+tcga}, Appendix \ref{app:pairwise-comparisons} Table \ref{table:pairwise-rumc+tcga}), confirming that enrichment significantly improved \texttt{Prost40M}, bringing its performance close to that of \texttt{Virchow2}, which remained the strongest encoder overall.
	
	\vspace{4mm}
	
	\noindent
	Interestingly, performance on the contemporary \texttt{UHC} cohort improved for all encoders after enrichment. Both \texttt{UHC} and \texttt{TCGA-PRAD} contain a higher proportion of intermediate-to-high ISUP grades than \texttt{RUMC}, \texttt{PLCO}, and \texttt{IMP} (Figure \ref{fig:isup}). Because ISUP grade is strongly associated with recurrence risk, training on \texttt{RUMC-only} exposes models to a grade distribution skewed toward lower grades, which can bias learned representations toward that part of the spectrum. Enrichment with \texttt{TCGA-PRAD} increased the representation of higher-grade cases during training, reducing this distribution shift and improving calibration across grades. Consistent with this interpretation, ISUP-stratified analyses on \texttt{UHC} showed that models trained on \texttt{RUMC-only} were skewed toward grade 2, whereas enrichment reduced this bias, improving performance in grades 3 and 5, and yielding a flatter performance profile across grades (Appendix \ref{app:subgroup-analysis-uhc} Figure \ref{fig:subgroup-analysis-uhc}). Approaches that explicitly balance grade representation during training, such as proportion-based sampling or loss weighting, may further mitigate this effect and represent a promising direction for future work.\\
	\\
	Overall, increasing training set heterogeneity yielded the largest gains for encoders with limited pretraining diversity, whereas models already pretrained on broad, heterogeneous corpora derived little additional benefit. Together with the results of Section \ref{sec:pretraining-diversity}, these findings indicate that robust generalization depends on an interaction between pretraining diversity and training data composition, and that aligning the clinical spectrum between training and target populations is critical for reliable prognostic modeling.
	
	\subsection{Deep learning adds prognostic value beyond CAPRA-S}
	\label{sec:res-add-value}
	
	Having established that deep learning models generalize across heterogeneous cohorts, we next asked whether features extracted by the deep learning model from routine prostatectomy histology provide prognostic value beyond established clinical risk models. Specifically, we evaluated whether a deep learning-derived risk score computed directly from digitized H\&E slides captures recurrence-associated signals that are complementary to CAPRA-S, a widely used clinicopathological risk score integrating surgical, pathological, and clinical variables. For reference, CAPRA-S alone achieved concordance indices of $0.725$ on \texttt{RUMC}, $0.761$ on \texttt{PLCO}, $0.772$ on \texttt{IMP}, and $0.747$ on \texttt{UHC} (Methods, Section \ref{sec:capra-s}). We then tested whether integrating image-derived predictions with CAPRA-S improves prognostic discrimination across independent cohorts.\\
	\\
	For each encoder, we derived a single patient-level deep learning risk score (DLRS) by averaging predictions across the five cross-validation folds. To test whether this image-derived score provides prognostic information beyond CAPRA-S, we fit multivariable Cox proportional hazards models including both CAPRA-S and the DLRS as covariates, and evaluated these models separately in each test cohort. We report: (i) the c-index of the DLRS alone, (ii) the c-index of the multivariable Cox model, (iii) the hazard ratio associated with the DLRS after adjustment for CAPRA-S, and (iv) the corresponding $p$-value from the Wald test. We focus here on models trained on the \texttt{RUMC+TCGA} splits, which showed superior discrimination (Table \ref{table:multivariable-cox-models-rumc-tcga}). For completeness, results for \texttt{RUMC-only} training are provided in Appendix \ref{app:joint-modeling-rumc} (Table \ref{table:multivariable-cox-models-rumc}).\\
	\\
	Across cohorts, integrating the DLRS with CAPRA-S consistently improved prognostic discrimination relative to either predictor alone. In most settings, the DLRS remained a statistically significant independent predictor after adjustment for CAPRA-S (hazard ratio $>1$, Wald $p<0.05$), demonstrating that the deep learning model captures recurrence-associated signals not fully accounted for by established clinical risk factors. While average performance varied modestly across encoders, with \texttt{Prost40M} achieving the highest mean joint c-index ($0.776$), followed closely by \texttt{Virchow2} ($0.769$) and \texttt{UNI} ($0.767$), the key finding was consistent across architectures: image-derived risk scores provided complementary prognostic information rather than acting as surrogates for CAPRA-S. Together, these results support the integration of deep learning-based histopathology biomarkers into multivariable risk models to improve postoperative risk stratification in prostate cancer.
	
	\begin{table}[htbp]
		\centering
		\caption{\textbf{Ensemble and joint modeling performance of models trained on \texttt{RUMC+TCGA}.} Summary of the prognostic performance of deep learning models evaluated on the four held-out test cohorts. For each cohort, the \emph{ensemble} c-index reflects the standalone performance of the deep learning model and is computed by averaging patient-level risk scores from the five \texttt{RUMC+TCGA} cross-validation folds. The \emph{joint} c-index is obtained by fitting a multivariable CoxPH model using both the ensemble risk score and the CAPRA-S score as covariates. For joint models, the hazard ratio (HR) with $95$\% confidence interval (CI) and the Wald test $p$-value indicate the independent contribution of the ensemble risk score after adjusting for CAPRA-S. Bold values indicate the highest c-index observed for each cohort.}
		\resizebox{\textwidth}{!}{
			\begin{tabular}{llccccc}
				\toprule
				\textbf{Cohort} & \textbf{Encoder} & \textbf{Ensemble} &  \textbf{Joint} & \textbf{HR (95\% CI)} & \textbf{p-value} \\
				\midrule
				\multirow{4}{*}{\texttt{RUMC}} & \texttt{Prost40M}    & 0.760           & \textbf{0.788}  & 1.80 (1.27-2.56)     & 0.001 \\
				& \texttt{UNI}         & 0.709           & 0.772           & 1.61 (1.12-2.29)     & 0.009 \\
				& \texttt{Virchow2}    & \textbf{0.767}  & 0.774           & 1.61 (1.14-2.26)     & 0.006 \\
				& \texttt{H-optimus-0} & 0.671           & 0.744           & 1.44 (1.01-2.06)     & 0.043 \\
				\midrule
				\multirow{4}{*}{\texttt{PLCO}} & \texttt{Prost40M}    & \textbf{0.711}  & \textbf{0.779}  & 1.32 (1.15-1.52)     & $<$ 0.001 \\
				& \texttt{UNI}         & 0.674           & 0.767           & 1.29 (1.12-1.49)     & $<$ 0.001 \\
				& \texttt{Virchow2}    & 0.675           & 0.765           & 1.25 (1.10-1.43)     & $<$ 0.001 \\
				& \texttt{H-optimus-0} & 0.651           & 0.770           & 1.17 (1.03-1.33)     & 0.016 \\                                                                                                                                                                
				\midrule                                                                                                                                                                                                                                                    
				\multirow{4}{*}{\texttt{IMP}}  & \texttt{Prost40M}    & 0.700           & \textbf{0.788}  & 1.21 (1.05-1.39)     & 0.008 \\                                                                                                                                 
				& \texttt{UNI}         & 0.669           & 0.779           & 1.07 (0.95-1.20)     & 0.260 \\                                                                                                                                                                
				& \texttt{Virchow2}    & \textbf{0.707}  & 0.779           & 1.02 (0.90-1.16)     & 0.754 \\                                                                                                                                                                
				& \texttt{H-optimus-0} & 0.563           & 0.775           & 0.89 (0.76-1.04)     & 0.154 \\                                                                                                                                                                
				\midrule                                                                                                                                                                                                                                                    
				\multirow{4}{*}{\texttt{UHC}}  & \texttt{Prost40M}    & 0.694           & 0.749           & 1.31 (1.07-1.62)     & 0.010 \\                                                                                                                                 
				& \texttt{UNI}         & 0.688           & 0.753           & 1.29 (1.08-1.53)     & 0.004 \\                                                                                                                                                                
				& \texttt{Virchow2}    & \textbf{0.735}  & \textbf{0.760}  & 1.52 (1.25-1.84)     & $<$ 0.001 \\                                                                                                                                                            
				& \texttt{H-optimus-0} & 0.686           & 0.752           & 1.23 (1.05-1.45)     & 0.013 \\                                                                                                                                                                
				\bottomrule                                                                                                                                                                                                                                                 
			\end{tabular}                                                                                                                                                                                                                                                
			\label{table:multivariable-cox-models-rumc-tcga}                                                                                                                                                                                                             
		}                                                                                                                                                                                                                                                             
	\end{table}                                                                                                                                                                                                                                                    
	
	\subsection{Model interpretability}                                                                                                                                                                                                                            
	
	Having established that our deep learning risk score provides prognostic information complementary to CAPRA-S, a critical question arises: \textit{which histological features actively drive the model’s predictions?} Answering this question is essential for understanding why histology-derived risk scores add value beyond established clinicopathological models. Doing so serves two distinct but equally important objectives. First, model interpretability allows clinical validation. By identifying the regions that most strongly influence predictions, we can evaluate whether the model relies on morphological features already known to correlate with prognosis, such as Gleason patterns \cite{Nguyen2024}. Demonstrating alignment with established pathology is crucial for earning clinician trust. Second, interpretability may uncover previously overlooked predictive cues. Because CAPRA-S does not account for within-grade morphology or subtle histopathological patterns, signals extracted directly from histology may explain the complementary prognostic information observed in Section \ref{sec:res-add-value}. Such complementary insights may ultimately support the discovery of new, human-interpretable biomarkers, motivating a systematic investigation into \textit{what} drives the model’s predictions.\\
	\\                                                                                                                                                                                                                                                             
	The term interpretability is widely used in computational pathology, yet it encompasses a range of distinct and sometimes conflicting definitions. In the context of outcome-driven models, we argue that interpretability is most meaningful when it reflects direct contribution to the model’s output. Under this definition, an interpretable feature is one whose presence or removal produces a measurable change in the predicted risk of biochemical recurrence, thereby directly linking tissue morphology to the model’s prognostic output.\\
	\\
	Our risk model is implemented as a Transformer-based multiple instance learning aggregator, which uses attention at multiple spatial scales to aggregate local tile-level representations into a contextualized patient-level prediction. Visualizing attention scores therefore provides a natural starting point for interpretability, as it reveals the regions where the model allocates representational focus in the whole-slide image. However, attention does not establish a causal relationship between these regions and the model’s output: it does not indicate whether a given region actively increases the predicted recurrence risk or instead provides evidence for lower risk. To address this limitation, we complement attention-based visualization with an occlusion-based perturbation analysis that directly quantifies the signed contribution of individual tiles to the model’s prediction, enabling a direct assessment of which histological patterns increase or decrease recurrence risk.
	
	\subsubsection{Attention as a first approximation of model interpretability}
	
\begin{figure*}[htbp]
		\centering
		\includegraphics[width=\textwidth]{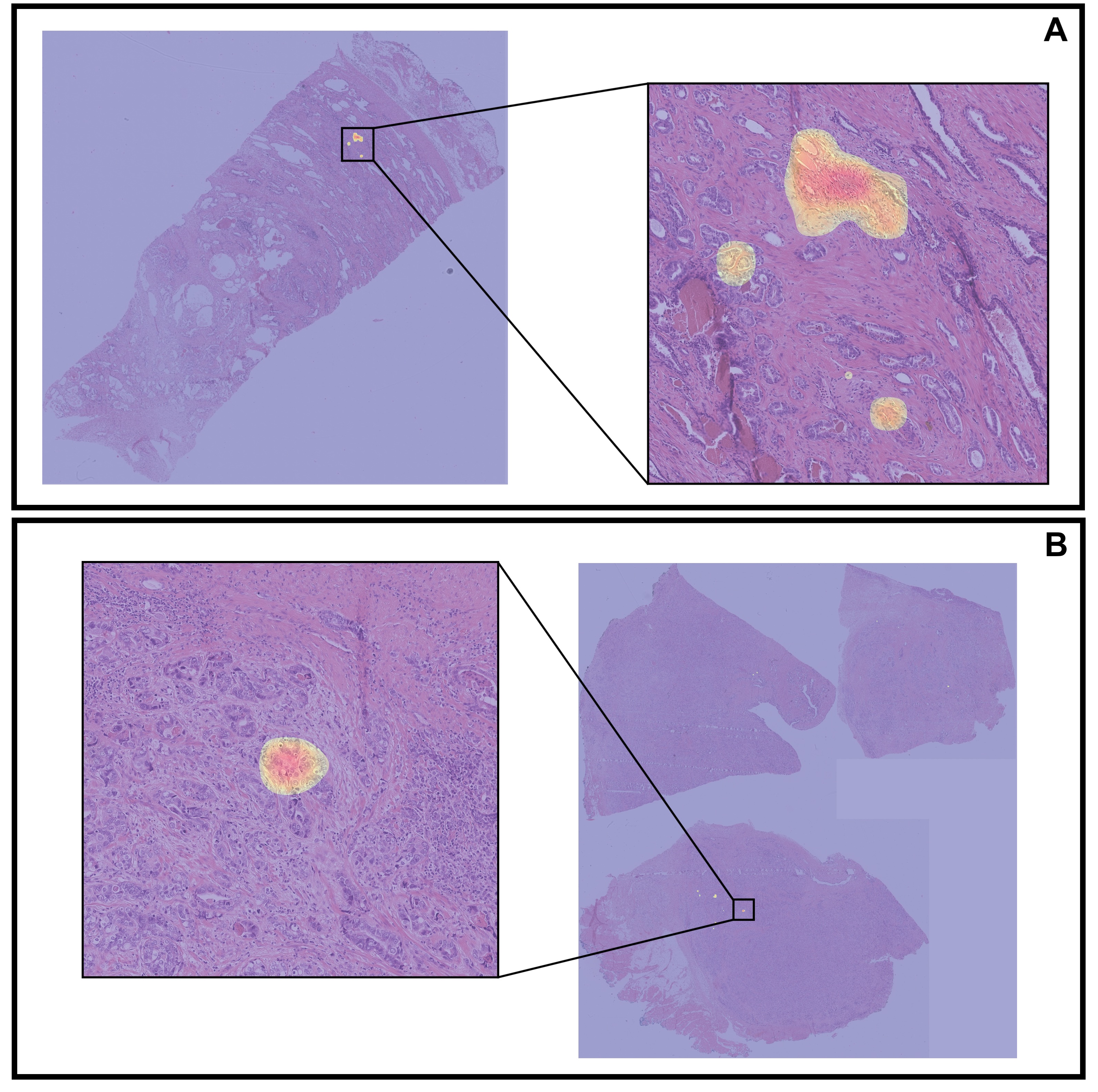}
		\caption{\textbf{Factorized attention maps reveal regions the model considers informative in prostatectomy slides.} Factorized attention heatmaps derived from the \texttt{Virchow2}-based model and overlaid on H\&E-stained slides. Each panel shows the whole-slide image with regions of high model attention, alongside a magnified view of the most salient region. \textbf{(A)} True positive case: a patient who experienced biochemical recurrence within a short time and was assigned a high risk score by the model. \textbf{(B)} True negative case: a patient who remained recurrence-free with long follow-up and was assigned a low risk score by the model.}
		\label{fig:attention-maps}
	\end{figure*}

	Transformer-based architectures derive much of their representational power from the self-attention mechanism, which allows the model to weight different regions of the input according to how informative they are for building internal representations. In the context of whole-slide images, attention maps provide a spatially resolved view of regions the model deems informative during inference. In Hierarchical Vision Transformers (HViTs), attention is computed at multiple spatial resolutions, with different Transformer blocks attending to complementary levels of tissue organization, ranging from local tile-level patterns to global slide-level context. Visualizing the attention map from any single block in isolation therefore provides only a partial view of how information is integrated across scales. To obtain a comprehensive, multi-scale view, we integrate attention scores across the hierarchical architecture by adapting the attention factorization framework of Grisi et al. \cite{Grisi2025}, aggregating attention multiplicatively across Transformer layers (Methods, Section \ref{sec:factorized-maps}). The resulting factorized attention maps highlight regions that receive consistent focus across scales.\\
	\\
	To illustrate how attention can support a qualitative understanding of regions the model considers informative, we visualize factorized attention maps for two representative patients in Figure \ref{fig:attention-maps}: one with early biochemical recurrence and a correctly predicted high-risk score (Figure \ref{fig:attention-maps}\textbf{A}), and one with long recurrence-free survival and a correctly predicted low-risk score (Figure \ref{fig:attention-maps}\textbf{B}). For clarity, attention maps are thresholded to display only the most salient regions, with magnified views highlighting prominent hotspots. These visualizations reveal spatial patterns consistently emphasized by the model and suggest histomorphological features potentially relevant for prognosis. However, attention alone does not indicate whether these regions increase or decrease the predicted recurrence risk. To move beyond localization and assess causal contribution, the next section turns to an outcome-grounded analysis.
	
	\subsubsection{Outcome-grounded interpretability via occlusion}
	
\begin{figure*}[htbp]
		\centering
		\includegraphics[width=\textwidth]{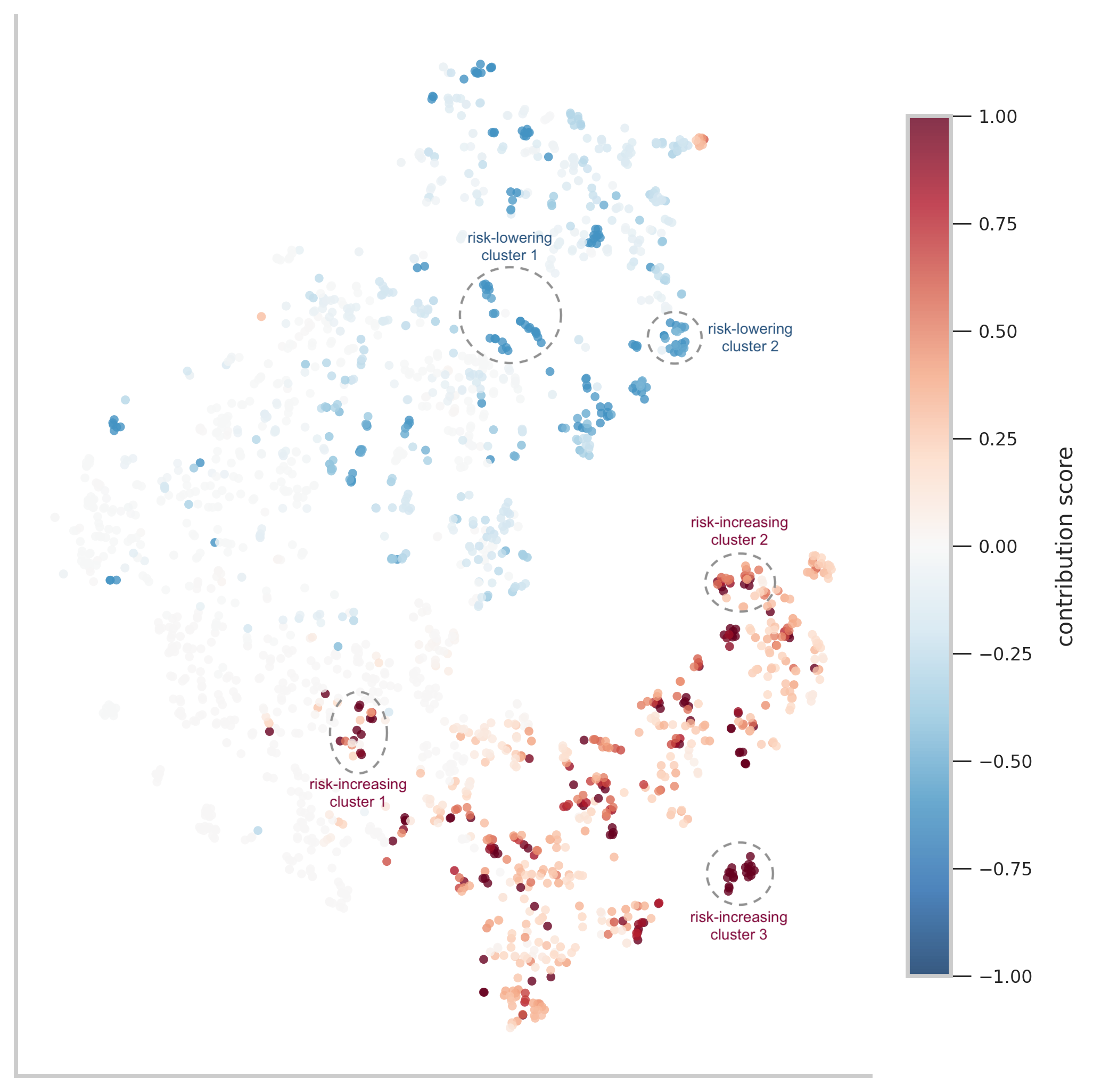}
		\caption{\textbf{t-SNE plot of model-refined tile representations reveals recurrence-associated morphology.} Across the \texttt{RUMC} test cohort, we selected tiles with the strongest positive or negative contributions to the recurrence risk estimated by the \texttt{Virchow2}-based model. Their model-refined embeddings are projected into two dimensions using t-SNE. Color encodes each tile’s signed contribution to the final prediction, with red indicating tiles associated with higher recurrence risk and blue indicating tiles associated with lower risk, highlighting gradients of prognostic relevance within clusters. Grey outlines denote the representative clusters subsequently reviewed by an expert pathologist.}
		\label{fig:t-SNE-rumc}
	\end{figure*}

	To operationalize our contribution-based definition of interpretability, we quantify how individual tissue regions influence predicted recurrence risk using an occlusion-based perturbation analysis (Methods, Section \ref{sec:occlusion-method}). For each patient, tiles are removed one at a time from the model’s input and the resulting change in predicted risk is measured. This procedure yields a signed contribution score for each tile, directly indicating whether it drives the prediction upward or downward and providing a causal attribution within the model’s decision process.\\
	\\
	To identify histological patterns that consistently affect model predictions, we focused on the most influential tiles from the held-out \texttt{RUMC} test set. For each patient, we selected the ten most risk-increasing and ten most risk-lowering tiles and examined whether shared morphology emerged across patients. To this end, we projected model-refined tile representations using t-SNE \cite{tsne}. We observed a clear separation between risk-increasing and risk-lowering tiles, with each group forming distinct clusters in the embedding space (Figure \ref{fig:t-SNE-rumc}). Notably, the magnitude of positive contributions was generally stronger than that of negative contributions, consistent with the clinical observation that benign morphology is shared across risk groups, while adverse features are primarily responsible for distinguishing high-risk patients.\\
	\\
	This analysis was deliberately restricted to the in-domain \texttt{RUMC} test set to ensure that clustering reflects biologically meaningful morphology rather than non-biological confounders such as staining variation or scanner-specific effects, which pathology foundation models are known to encode strongly \cite{dejong2025}. By operating on the risk model's internal representations instead of frozen foundation model embeddings, we aimed to further mitigate center-driven batch effects and focus on morphology relevant to recurrence prediction.
	
	\vspace{4mm}
	\noindent
	An expert pathologist (E.M.) reviewed representative clusters of risk-increasing and risk-lowering tiles to characterize the underlying morphology (Appendix \ref{app:cluster-analysis}). Risk-increasing clusters frequently captured adverse histologies, including large cribriform glands (Figure \ref{fig:tsne_high_cluster1}), fused Gleason pattern 4 glands, and single-cell pattern 5 carcinoma (Figure \ref{fig:tsne_high_cluster2}), whereas risk-lowering clusters were enriched for well-formed glands or benign tissue (Figure \ref{fig:tsne_low_cluster1}). This alignment with established prognostic signals supports the biological and clinical validity of the model.\\
	\\
	Beyond tumor-centric features, one risk-increasing cluster contained little or no carcinoma and was dominated by fibromuscular stroma or visually heterogeneous non-tumor tissue (Figure \ref{fig:tsne_high_cluster3}). Its association with increased recurrence risk suggests that the model captures prognostic cues beyond tumor morphology -- signals that current risk models do not account for. Similarly, the presence of risk-lowering clusters containing specific Gleason pattern 4 morphologies (Figure \ref{fig:tsne_low_cluster2}) highlights within-grade heterogeneity that is not distinguished by CAPRA-S. Together, these findings provide biological grounding for the independent risk signal identified in Section \ref{sec:res-add-value}. Comprehensive investigation of these signals is an important direction for future work but lies beyond the scope of this study.\\
	\\
	This outcome-grounded analysis provides a general blueprint for causal interpretation of pathology prognostic models. Together, our findings reveal that the model integrates both well-established adverse tumor morphology and additional prognostic features not routinely assessed in clinical practice. This duality underscores its alignment with expert pathology while highlighting deep learning’s potential to uncover novel biomarkers and broaden our understanding of the mechanisms underlying biochemical recurrence.
	
	\section{Discussion}
	
	In this study, we developed and validated a deep learning-based biomarker for predicting biochemical recurrence directly from routine H\&E-stained prostatectomy specimens, trained end-to-end on time-to-event outcomes and evaluated across multiple independent cohorts. Beyond demonstrating strong prognostic performance, our results provide broader insights into how computational pathology models can be trained, evaluated, and interpreted to support clinically meaningful risk stratification.\\
	\\	
	First, our results indicate that robust generalization in histology-based prognostic modeling depends on an interaction between pretraining diversity and downstream training data composition, rather than on either factor in isolation. When downstream training data were drawn from a single institution, models pretrained on large, heterogeneous corpora generalized substantially better than narrowly pretrained, organ-specific encoders. Conversely, enriching downstream training with multi-institutional cases primarily benefited encoders with limited pretraining diversity, while providing little additional gain for foundation models. Beyond variability in data acquisition protocols, our results show that differences in the underlying patient mix also shape generalization. Performance on the \texttt{UHC} cohort improved for all encoders after adding \texttt{TCGA-PRAD} to the training set, consistent with these cohorts sharing a similar clinical spectrum distinct from \texttt{RUMC}, \texttt{PLCO}, and \texttt{IMP}. Together, these results point to a general principle: pretraining diversity and downstream data composition play complementary roles in shaping model generalization, with their relative importance depending on where heterogeneity is introduced in the pipeline. While broad pretraining can mitigate limited downstream data diversity, aligning the clinical spectrum between training and deployment settings remains critical for all models. This insight is particularly relevant as pathology foundation models are increasingly applied across institutions with heterogeneous patient populations, scanners, and clinical practices.\\
	\\
	Second, our study provides a mechanistic explanation for why deep learning-derived histology scores add prognostic value beyond established clinical risk models such as CAPRA-S. Through outcome-grounded interpretability analyses, we show that the model captures both well-established adverse tumor morphologies and additional recurrence-associated features, including within-grade architectural heterogeneity and non-tumor or stromal patterns. These features may encode biologically relevant interactions that are not explicitly captured in current clinicopathological risk models relying on coarse categorical grades, providing a biological rationale for the independent prognostic signal observed when combining the deep learning risk score with CAPRA-S. More broadly, our findings illustrate how deep learning applied to histopathology can reveal prognostic signals that may not be readily accessible through routine human assessment. By learning from large numbers of cases and operating across multiple spatial scales, the model can integrate subtle and spatially distributed cues that are difficult to conceptualize or quantify systematically in clinical practice.\\
	\\	
	Third, our results have practical implications for deploying pathology AI systems in real-world clinical settings. We demonstrate consistent performance across cohorts spanning diverse institutions, scanners, patient populations, and definitions of biochemical recurrence. Generic postoperative risk models are inherently limited: patients with identical CAPRA-S scores can experience markedly different disease trajectories. Moving toward precision oncology therefore requires incorporating additional patient-specific information into risk stratification. Routine histopathology encodes rich, spatially resolved information about tumor architecture, grade heterogeneity, and tumor-stroma context that is systematically available for nearly all patients but remains largely unquantified in current risk models. Deep learning offers a principled framework to extract and integrate these signals into patient-level predictions at scale. While future progress will likely involve multimodal models integrating imaging, genomics, and clinical data, building robust and well-validated unimodal histology models is a critical and necessary step toward that goal.\\
	\\
	This work also has limitations that point to important directions for future research. Although pathology foundation models improved robustness, their performance was not uniformly stable across all settings, indicating that large-scale pretraining alone does not eliminate susceptibility to cohort-specific biases. More systematic strategies for balancing clinically relevant variables during training may further improve generalization. In addition, while occlusion-based interpretability provides a more faithful assessment of tile-level contributions than attention alone, rigorous validation of model-derived patterns -- such as blinded, multi-expert review across cohorts -- will be essential to translate these insights into reproducible pathological concepts. Finally, prospective studies will be required to assess how histology-based risk scores influence clinical decision-making and patient outcomes when integrated into real-world workflows.\\
	\\
	In summary, our findings highlight a broader role for histology-based AI in precision oncology. Deep learning models trained directly on routine prostatectomy slides can uncover prognostic signals that complement established clinical risk scores and capture aspects of tumor biology not currently used in postoperative care. By enabling systematic, fine-grained characterization of tumor morphology, computational pathology offers a path toward more individualized risk stratification and treatment planning in prostate cancer.
	
	\section{Methods}
	
	\subsection{Introducing a small-scale, prostatectomy-specific encoder}
	\label{sec:dino-vit-s}
	
	While pathology foundation models are pretrained on extremely large and diverse datasets, their generic features, though effective across many organs and diseases, may lack sensitivity to the subtle morphological patterns of a specific organ. To evaluate the potential benefits of organ-specific pretraining, we developed \texttt{Prost40M}, a ViT-Small pretrained with DINO \cite{Caron2021}.\\
	\\
	In DINO, a student network $g_{\theta_s}$ is trained to match the output of a teacher network $g_{\theta_t}$ under different augmented crops of the same image, encouraging consistency across local and global contexts. Given a $(256, 256)$ pixel image $\mathbf{x}$, a set $V$ of different views is generated: $8$ local crops of $(96, 96)$ pixels and $2$ global crops of $(224, 224)$ pixels. All views are passed through the student, while only the global views are passed through the teacher. Given a pair of input views, both networks output probability distributions over $K$ dimensions, denoted $p_s$ and $p_t$, respectively. These distributions are matched by minimizing the cross-entropy loss with respect to the student parameters $\theta_{s}$:
	
	\begin{equation*}
		\min _{\theta_{s}} \sum_{\mathbf{v} \in \left\{\mathbf{v}^g_1, \mathbf{v}^g_2\right\}} \sum_{\substack{\mathbf{v'} \in V \\ \mathbf{v'} \neq \mathbf{v}}}  H\left(p_{t}\left(\mathbf{v} \right), p_{s}\left(\mathbf{v'}\right)\right)
	\end{equation*}\\
	\\
	\vspace{-1.4cm}\\
	\\
	where $H(a, b)=-a \log b$.\\
	\\
	\texttt{Prost40M} was trained on approximately 40 million tiles extracted from 1888 H\&E-stained prostatectomy slides -- 449 sourced from the 403 patients in the \texttt{TCGA-PRAD} cohort and 1439 sourced from the 508 patients in the \texttt{RUMC} development set. Using the \texttt{slide2vec} library, we extracted non-overlapping $(256, 256)$ pixel tiles at 0.50 mpp, discarding any with less than 25\% tissue coverage. We pretrained \texttt{Prost40M} for a single epoch, as additional pretraining epochs yield only marginal gains due to the high visual redundancy between histology tiles.
	
	\subsection{Discrete modeling of biochemical recurrence}
	\label{sec:discrete-time-modelling}
	
	Biochemical recurrence prediction is an ordinal regression task that models time-to-event distributions, where the event (recurrence) is not always observed and may be right-censored. Patients with observed BCR were treated as events, and patients without observed BCR were right‑censored at their last contact.\\
	\\
	Let $T$ be a continuous random variable denoting the time to biochemical recurrence. The recurrence-free survival function $f_{\text{bcr}}(T > t \mid X)$ gives the probability that a patient with covariates $X$ remains recurrence-free beyond time $t$, and the hazard function $f_{\text{hazard}}(T = t \mid T \geq t, X)$ denotes the probability of BCR at time $t$, defined as:
	
	\begin{equation*}
		f_{\text{hazard}}(T = t) = \lim_{\Delta t \to 0} \frac{P(t \leq T < t + \Delta t \mid T \geq t)}{\Delta t}
	\end{equation*}
	\vspace{-4mm}\\
	\\
	Traditionally, a Cox Proportional Hazards (CoxPH) model is used to estimate the hazard function, and in deep learning this is implemented by optimizing the Cox partial log-likelihood. However, this loss depends on pairwise comparisons between patients within a batch, making it unsuitable for whole-slide image training where memory constraints and variable number of tiles make training with large batch sizes impractical. To address this, we adopt the discrete-time survival modeling from Chen et al. \cite{chen2021multimodal}, which allows training with a batch size of one. Specifically, we partition the continuous time scale into four non-overlapping bins: $[t_0, t_1)$, $[t_1, t_2)$, $[t_2, t_3)$, and $[t_3, t_4)$, defined by the quartiles of observed recurrence times among uncensored patients. Each patient is then assigned a discrete label $Y_i \in \{0, 1, 2, 3\}$ indicating the interval in which recurrence was observed, or in which censoring occurred if no event was recorded. Given a patient-level feature vector $h_i$, the model predicts the discrete hazard for each interval. The hazard function for bin $b$ is defined as:
	
	\begin{equation*}
		f_{\text{hazard}}(b \mid h_i) = P(T = b \mid T \geq b, h_i)
	\end{equation*}\vspace{-4mm}\\
	\\
	which relates to the recurrence-free survival function:
	
	\begin{equation*}
		f_{\text{bcr}}(b \mid h_i) = \prod_{u=1}^{b} \left(1 - f_{\text{hazard}}(u \mid h_i)\right)
	\end{equation*}\vspace{-4mm}\\
	\\
	We train the model using the log-likelihood function for a discrete survival model \cite{Zadeh2021}, accounting for censorship:
	
	\begin{align*}
		\mathcal{L} =\ & - c_i \cdot \log  \left( f_{\text{bcr}}(Y_i \mid h_i) \right) \nonumber \\
		& - (1 - c_i) \cdot \log \left( f_{\text{bcr}}(Y_i - 1 \mid h_i) \right) \nonumber \\
		& - (1 - c_i) \cdot \log \left( f_{\text{hazard}}(Y_i \mid h_i) \right)
	\end{align*}\vspace{-4mm}\\
	\\
	where $c_i = 1$ for censored patients and $0$ otherwise. Following prior work \cite{chen2021multimodal}, we up-weight the contribution of uncensored patients during training, leading to the final objective:
	
	\begin{equation}
		\label{equation:loss}
		\mathcal{L}_{\text{bcr}} = (1 - \alpha) \cdot \mathcal{L} + \alpha \cdot \mathcal{L}_{\text{uncensored}}
	\end{equation}\vspace{-4mm}\\
	\\
	where $\mathcal{L}_{\text{uncensored}}$ corresponds to the loss term for uncensored patients:
	
	\begin{align*}
		\mathcal{L}_{\text{uncensored}} = & - (1 - c_i) \cdot \log \left( f_{\text{bcr}}(Y_i - 1 \mid h_i) \right) \nonumber \\
		& - (1 - c_i) \cdot \log \left( f_{\text{hazard}}(Y_i \mid h_i) \right)
	\end{align*}\vspace{-4mm}\\
	\\
	This formulation enables stable training on large-scale WSIs using a batch size of one, without relying on inter-sample risk comparisons.
	
	\subsection{Weakly supervised prediction of BCR with Hierarchical Vision Transformers}
	\label{sec:hipt}
	
	Whole-slide images exhibit a natural hierarchical structure, with information distributed across multiple scales. At the finest level, cell-centric regions reveal nuclear morphology and subcellular detail. At intermediate scales, patterns of cell-to-cell interaction become apparent. At progressively larger scales, broader tissue architecture emerges, capturing macro-level organization and interactions between cellular clusters. Drawing inspiration from this layered structure, Chen et al. \cite{chen2022} introduced the Hierarchical Image Pyramid Transformer (HIPT), a three-stage model that aggregates information at progressively coarser spatial resolutions. This multi-scale modeling is essential for capturing the complexity of intra-tumoral heterogeneity observed across entire slides.\\
	\\
	To predict biochemical recurrence from whole-slide images, we adopt a weakly supervised multiple instance learning framework largely based on the HIPT. Its multiscale approach is particularly suited to biochemical recurrence prediction: fine-grained nuclear morphology may reveal subtle indicators of tumor aggressiveness, whereas larger-scale tissue architecture reflects growth patterns and tumor-stroma interactions. By jointly modeling these complementary scales, HIPT learns rich, context-aware slide representations and has demonstrated state-of-the-art performance in tasks such as cancer subtyping and prostate cancer grading \cite{chen2022,Grisi2025}.\\
	\\
	Because of GPU memory constraints, the model cannot be trained end-to-end. We therefore decouple feature extraction from feature aggregation. In the feature extraction stage, each slide is divided into non-overlapping $2048$ pixel regions at $0.50$ mpp, which are further unrolled into $(256,256)$ tiles. Each tile is passed through a pretrained encoder, producing region-level features of shape $(64, d)$, where $d$ is the encoder's embedding dimension. Stacking all region embeddings from a case yields a patient-level feature $h_i$ of shape $(M_{i,2048}, 64, d)$, where $M_{i,2048}$ denotes the number of $2048$ pixel regions extracted for patient $i$. In the feature aggregation stage, a region-level Transformer compresses each $(64, d)$ region feature into a 192-dimensional embedding. The resulting sequence of shape $(M_{i,2048}, 192)$ is processed by a slide-level Transformer, which pools across regions to form a single 192-dimensional whole-slide embedding. A fully connected layer finally projects this embedding to discrete recurrence labels. All aggregator components are jointly trained using the discrete-time survival loss defined in Equation \ref{equation:loss} (Section \ref{sec:discrete-time-modelling}). We evaluated three recent pathology foundation models as tile encoders, chosen for their strong performance in the HEST-Benchmark \cite{jaume2024}:
	
	\begin{itemize}
		\item \texttt{UNI} \cite{Chen2024} is a ViT-L/$16$ model with $307$ million parameters trained on over $100$ thousand WSIs using DINOv2.
		\item \texttt{Virchow2} \cite{zimmermann2024} is a ViT-H/$14$ model with $632$ million parameters trained on $3.1$ million WSIs using a modified DINOv2 framework tailored for histopathology.
		\item \texttt{H-optimus-0} \cite{hoptimus0} is a ViT-g/$14$ model with $1.1$ billion parameters trained on a proprietary collection of more than $500$ thousand WSIs using DINOv2.
	\end{itemize}
	
	\noindent
	In all experiments, aggregator components are optimized for up to 100 epochs using Adam \cite{kingma2017} with a base learning rate of $2 \times 10^{-4}$ (halved every 20 epochs) and weight decay of $10^{-5}$. We apply early stopping after 20 consecutive epochs without improvement on the tuning loss, with a minimum of 50 epochs completed. We use a batch size of 1 with 32 gradient accumulation steps.
	
	\subsection{Factorized attention maps}
	\label{sec:factorized-maps}
	
	To factorize attention scores across the entire hierarchical architecture, we build on the work of Grisi et al. \cite{Grisi2025}, replacing summation with multiplication across transformers. Let $N$ be the total number of Transformers involved in the pretraining or the training processes. Let $n$ be the number of frozen Transformers. For the pixel with $(x,y)$ coordinates in the slide, we denote by $a^j_{(x,y)}$ the attention score of the last self-attention layer of the $j$-th Transformer $\text{T}_j$. We compute the factorized attention score for that pixel, $a_{(x,y)}$, as:
	
	\begin{equation} \label{eq:factorized-attention}
		a_{(x,y)} = \prod_{j=0}^{N-1} a^j_{(x,y)} \left[ \gamma (1-\mathbbm{1}_{\text{F}}(\text{T}_j)) + (1-\gamma) \mathbbm{1}_{\text{F}}(\text{T}_j) \right]
	\end{equation}
	\\
	\noindent
	with $\mathbbm{1}_{\text{F}}(\text{T}_j)$ the indicator function defined on the set of Transformers $\left\{\text{T}_j\right\}_{j=0}^{N-1}$ and equal to $1$ if $\text{T}_j$ is frozen, $0$ otherwise. In this study, we set $\gamma = 1$ in Equation \ref{eq:factorized-attention} to nullify the contribution of the frozen encoder and focus on the attention from the region-level and slide-level Transformers, which are explicitly trained on the prediction task.\\
	\\
	Attention scores were normalized at the slide level, mapped to RGB colors using a diverging colormap, and overlaid on top of their respective spatial locations in the whole-slide image. To facilitate interpretation, scores were smoothed with a Gaussian kernel and values below $0.5$ were filtered out, leaving only salient regions visible. Hotspots of high attention, shown in red, indicate positive evidence with strong relative contribution to the model prediction compared to other regions. Heatmaps were overlaid on the original H\&E slide with a transparency value of $0.5$ to simultaneously visualize the underlying tissue morphology.
	
	\subsection{Tile-level contribution quantification via occlusion}
	\label{sec:occlusion-method}
	
	We assessed the influence of individual tiles on the model’s predictions using an occlusion-based perturbation approach. For each patient $i$, let $\text{S}_i$ denote the sequence of feature vectors produced by a pretrained encoder for all non-overlapping $(256, 256)$ pixel tiles sampled at 0.50 mpp in that case:
	
	\begin{equation*}
		\text{S}_i = (x^i_{1}, x^i_{2}, \dots, x^i_{n_i})
	\end{equation*}\vspace{-4mm}\\
	\\
	where $n_i$ is the number of tiles for case $i$. The trained recurrence-risk model is written as:
	
	\begin{equation*}
		\phi : (x^i_{1}, x^i_{2}, \dots, x^i_{n_i}) \mapsto r_i \in \mathbb{R}
	\end{equation*}\vspace{-4mm}\\
	\\
	We first computed the patient-level risk score:
	
	\begin{equation*}
		r_i^{\mathrm{legacy}} = \phi \bigl(\text{S}_i\bigr)
	\end{equation*}\vspace{-4mm}\\
	\\
	To assess the influence of tile $t$ on the model's prediction, we constructed an occluded version of the case by removing that tile from the sequence, resulting in a sequence of length $n_i-1$:
	
	\begin{equation*}
		\text{S}_i^{t} = (x^i_{1}, \dots, x^i_{t-1}, x^i_{t+1}, \dots, x^i_{n_i})
	\end{equation*}\vspace{-4mm}\\
	\\
	Thus $\text{S}_i^{t}$ contains one fewer element than $\text{S}_i$. The corresponding occluded risk score is:
	
	\begin{equation*}
		r_i^{t} = \phi \bigl(\text{S}_i^{t}\bigr)
	\end{equation*}
	\vspace{-4mm}\\
	\\
	The contribution score of tile $t$ in case $i$ was defined as:
	\begin{equation*}
		c_i^t = r_i^{\mathrm{legacy}} - r_i^{t}
	\end{equation*}\vspace{-4mm}\\
	\\
	By construction, $c_i^t > 0$ indicates a \emph{risk-increasing} tile (occlusion decreases predicted risk), whereas $c_i^t < 0$ indicates a \emph{risk-lowering} tile (occlusion increases predicted risk). For each case $i$, we retained the $K = 10$ tiles with the most positive contributions and the $K = 10$ tiles with the most negative contributions. To mitigate outliers, contribution scores were clipped at the 5th ($q_{5}$) and 95th ($q_{95}$) percentiles computed across all tiles from all 100 cases in the \texttt{RUMC} test set:
	
	\begin{equation*}
		\hat{c}_i^t = \min \bigl(\max(c_i^t, q_{5}),\, q_{95}\bigr)
	\end{equation*}\vspace{-4mm}\\
	\\
	Clipped values were normalized by the maximum absolute contribution score, yielding normalized scores in $[-1,1]$:
	
	\begin{equation*}
		\bar{c}_i^t = \frac{\hat{c}_i^t}{\max_{j,k} |\hat{c}_j^{k}|}
	\end{equation*}\vspace{-4mm}\\
	\\
	For each selected tile, we extracted the model-refined representation, corresponding to the output of the final region-level transformer layer prior to attention pooling. These embeddings were projected into two dimensions using t-SNE, with points colored by their normalized contribution $\bar{c}_i^t$. Within five visually coherent clusters, we randomly sampled up to 20 representative tiles for qualitative assessment by an expert pathologist (Appendix \ref{app:cluster-analysis}).
	
	\subsection{Statistics}
	\label{sec:statistics}
	
	We assessed statistical significance of performance differences using nonparametric resampling methods. All analyses were conducted at the patient level using ensemble-derived risk scores. Model performance was quantified using Harrell’s concordance index (c-index), with 95\% confidence intervals estimated by event-stratified bootstrapping with $4000$ resamples. We report $p$-values and adjusted $q$-values in Appendix \ref{app:statistics}.
	
	\paragraph{Pairwise model comparisons.} To compare encoders within each test cohort, we performed paired bootstrap resampling of the difference in c-index between two models, using two-sided tests. Resulting $p$-values were adjusted for multiple comparisons with the Benjamini-Hochberg procedure.
	
	\paragraph{Training set enrichment effect.} To evaluate the effect of enriching the training set with \texttt{TCGA-PRAD} cases, we applied the same procedure to compare ensemble risk scores from models trained on \texttt{RUMC-only} versus \texttt{RUMC+TCGA}, stratified by encoder and test cohort.
	
	\subsection{Datasets}
	\label{sec:materials}
	
	We conducted large-scale evaluation across five multi-institutional radical prostatectomy cohorts. All datasets included digitized whole-slide images at approximately $0.25$ microns per pixel and clinical follow-up for biochemical recurrence, providing time-to-event or time-to-censoring information. Additional dataset details, including inclusion and exclusion criteria and scanner specifications, are provided in Appendix \ref{app:datasets}.
	
	\subsubsection{RUMC}
	
	The Radboud University Medical Center (\texttt{RUMC}) cohort includes $608$ patients who underwent radical prostatectomy between $1992$ and $2012$ in Nijmegen, The Netherlands. For each patient, histopathology slides corresponding to the highest-grade tumor -- identified from the original pathology report -- were retrieved from the hospital archive and digitized using a 3DHistech P1000 scanner. This cohort was part of the \texttt{LEOPARD} challenge (LEarning biOchemical Prostate cAncer Recurrence from histopathology sliDes). We used the $508$ publicly released cases for model development and $100$ held-out test cases, totaling $1723$ slides. Clinico-pathological variables included preoperative PSA, extracapsular extension, surgical margin status, seminal vesicle invasion, lymph node involvement, and prostatectomy Gleason grade. Gleason grading for this cohort was based on the original pathology report and was not re-evaluated.
	
	\subsubsection{TCGA-PRAD}
	
	The Cancer Genome Atlas Prostate Adenocarcinoma (\texttt{TCGA-PRAD}) cohort consists of $403$ patients with at least one associated prostatectomy diagnostic slide. Although TCGA provides annotations for BCR, we found these labels to be inconsistent across different clinical files. To construct a reliable ground truth for our study, we curated BCR status and time-to-event information by parsing the case-specific \texttt{.xml} files available through the TCGA portal. We identified $298$ patients with BCR information ($62$ events), while the remaining $105$ patients were unlabeled and used only for self-supervised pretraining (Section \ref{sec:dino-vit-s}), for a total of $449$ slides.
	
	\subsubsection{PLCO}
	
	The Prostate, Lung, Colorectal, and Ovarian (\texttt{PLCO}) Cancer Screening Trial enrolled approximately $155,000$ participants between $1993$ and $2001$, with prostate cancer follow-up available through December $31$, $2009$. For this study, we selected prostate cancer patients whose primary treatment was radical prostatectomy. An experienced pathologist (J.B.) manually reviewed all available histopathology slides to identify those containing cancer, allowing us to exclude benign slides from further analysis. In the end, we included $723$ patients, for a total of $1971$ slides. BCR status was determined from standardized PSA follow-up records, and all cases were regraded according to the latest ISUP guidelines \cite{Epstein2021} by pathology residents under supervision of expert uropathologists. These cases were also part of the \texttt{LEOPARD} challenge.
	
	\subsubsection{IMP}
	
	The Instituto Mário Penna (\texttt{IMP}) cohort comprised $421$ patients who underwent radical prostatectomy in $2016$ in Belo Horizonte, Brazil. Patients were all followed up until $2022$. Histopathology slides for each case were digitized using a Motic Easy Scan Infinity $60$N scanner, resulting in a total of $2091$ digital prostatectomy slides. Patient-level clinico-pathological data included preoperative PSA levels, prostatectomy Gleason grades, extracapsular extension, surgical margin status, seminal vesicle invasion, and lymph node involvement.
	
	\subsubsection{UHC}
	
	The University Hospital Cologne (\texttt{UHC}) cohort included $330$ patients who underwent radical prostatectomy between $2015$ and $2020$ in Cologne, Germany. Patients were followed through $2023$ in accordance with European Association of Urology (EAU) guidelines. Histopathology slides for each case were digitized using a Hamamatsu NanoZoomer S$360$ scanner, resulting in a total of $1028$ digital prostatectomy slides. Patient-level clinico-pathological data included preoperative PSA, prostatectomy Gleason grades, and pathological staging.
	
	\begin{figure}[!h]
		\centering
		\includegraphics[width=\textwidth]{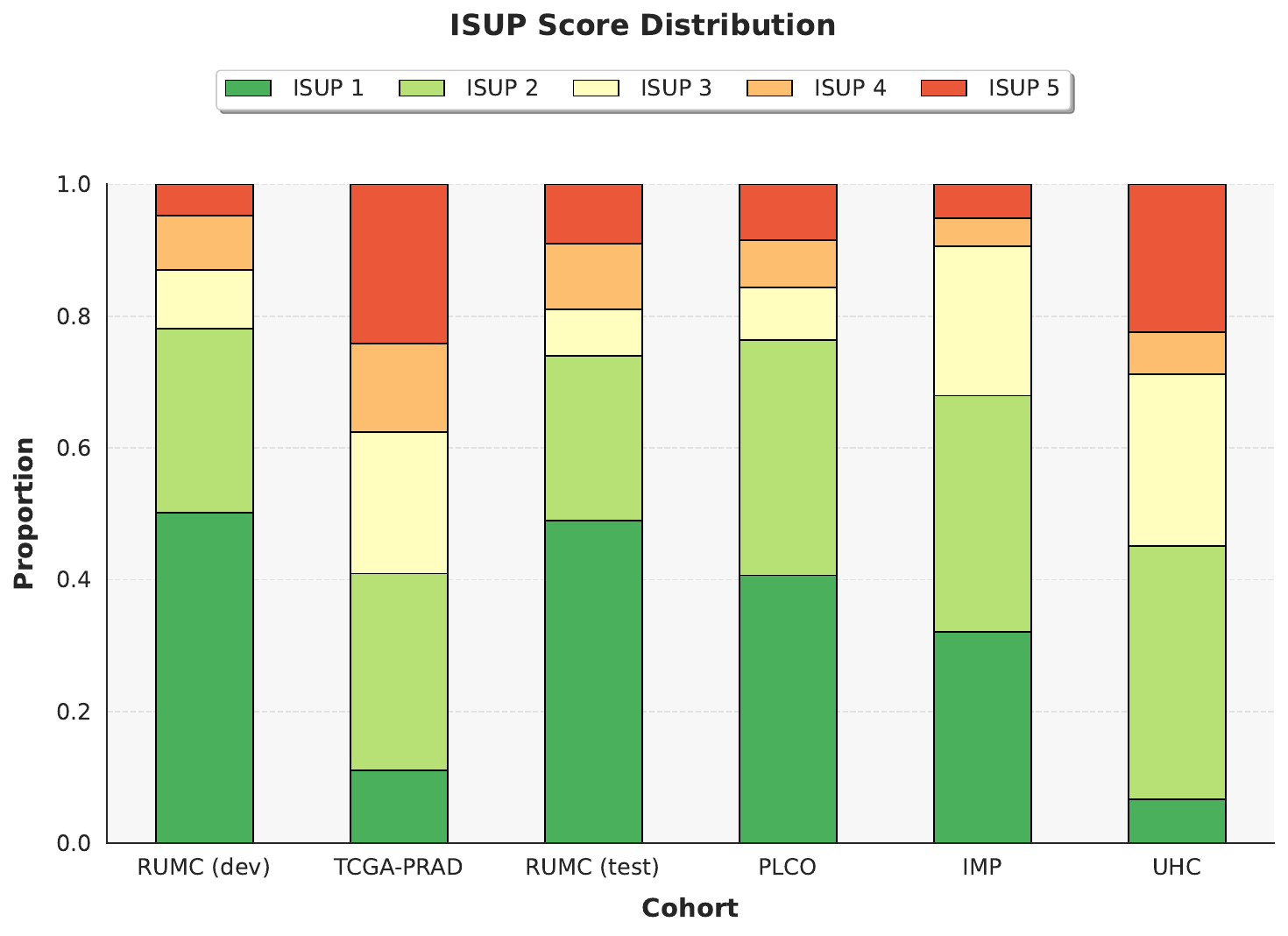}
		\caption{\textbf{ISUP grade distribution across cohorts.} Stacked bar plots show the proportion of patients in each cohort used in this study, grouped by ISUP grade. Each bar represents a cohort, with colors indicating ISUP grades 1 (green) to 5 (red). Differences in grade distributions reflect variation in recruitment periods, inclusion criteria, and treatment guidelines.}
		\label{fig:isup}
	\end{figure}
	
	\subsection{CAPRA-S}
	\label{sec:capra-s}
	
	CAPRA-S combines six clinicopathological variables into a $0$--$12$ point scale (Table \ref{tab:capra-s}) to stratify patients into three groups at low $(0--2)$, intermediate $(3--5)$, and high ($\geq 6$) risk of recurrence \cite{Cooperberg2011}.
	
	\begin{table}[t]
		\centering
		\caption{Scoring system used for calculating the CAPRA-S score following radical prostatectomy. Each variable contributes a fixed number of points: preoperative prostate-specific antigen (PSA) level in ng/mL, prostatectomy Gleason score (pGS), status of the surgical margin (SM), seminal vesicle invasion (SVI), extracapsular extension (ECE), and lymph node involvement (LNI). Points are summed to yield the CAPRA-S score.}
		\label{tab:capra-s}
		\begin{tabular}{lllllll}
			\toprule
			Variable & Value & Points && Variable & Value & Points \\
			\midrule
			
			\textbf{PSA} & 0--6 & 0 && \textbf{pGS} & 2--6 & 0 \\
			& 6.01--10 & 1 &&                    & 3+4 & 1 \\
			& 10.01--20 & 2 &&                   & 4+3 & 2 \\
			& $>$20 & 3 &&                       & 8--10 & 3 \\
			
			\addlinespace
			\textbf{SM} & Negative & 0 && \textbf{ECE} & No & 0 \\
			& Positive & 2 &&              & Yes & 1 \\
			
			\addlinespace
			\textbf{SVI} & No & 0 && \textbf{LNI} & No & 0 \\
			& Yes & 2 &&              & Yes & 1 \\
			
			\bottomrule
		\end{tabular}
	\end{table}
	
	\vspace{4mm}
	
	\noindent
	Several of these clinicopathological variables were missing or partially available in some cohorts. When possible, we inferred surrogate values using information derived from pathological staging, based on the American Joint Committee on Cancer 8th edition TNM system for prostate cancer, which is routinely used in clinical and pathological assessment. Specifically:	
	
	\begin{itemize}[itemsep=0pt, parsep=0pt]
		\item Extracapsular extension: assumed for pT-stage $\geq$ T3a
		\item Seminal vesicle invasion: assumed for pT-stage $\geq$ T3b
		\item Surgical margin positivity: assumed for pT-stage $\geq$ T3
		\item Lymph node involvement: based on pN1 status
	\end{itemize}
	
	\noindent
	These assumptions applied to \texttt{RUMC}, \texttt{UHC}, and \texttt{IMP}. For \texttt{PLCO}, pathological T staging was based on earlier AJCC editions: the 4th edition in 22\% of cases and the 5th edition in the remaining 78\%. In this cohort, the only deviation from the above assumptions concerned seminal vesicle invasion, which was assumed for pT-stage $\geq$ T3c under the AJCC 4th edition, rather than $\geq$ T3b. No change was required for cases staged according to the AJCC 5th edition. Missing values were handled via median imputation within each cohort.
	
	\subsection{Data preprocessing and experimental setup}
	\label{sec:preprocessing}
	
	\paragraph{Preprocessing.} To prepare the data for model training, we first apply a deep learning-based tissue segmentation model to isolate tissue from the background in each whole-slide image. For patients where multiple slides are available, the segmented tissue regions are stitched into a unified, larger slide to enable patient-level analysis in a unified coordinate frame (Figure \ref{fig:pipeline}\textbf{A}) using a tissue-packing algorithm \cite{packing}. Slides from the \texttt{PLCO} dataset contain pen markings, which we detect and remove from the tissue masks using an artifact segmentation model \cite{lucassen2024}. We then use the Python library \texttt{slide2vec} to extract non-overlapping $2048$ pixel regions at a resolution of $0.50$ mpp. To account for slight variations in resolution introduced by different scanners, \texttt{slide2vec} uses a spacing tolerance parameter $\tau$, which defines how much the native spacing can deviate from the target spacing during tiling. We set $\tau$ to 5\% to accept native spacings between $0.475$ and $0.525$ mpp. If no native spacing falls within this range, a larger region is extracted at a higher resolution and resized to approximate $0.50$ mpp. We filter out low-content regions by retaining only those with at least $1\%$ tissue coverage, as estimated by the segmentation mask (Figure \ref{fig:pipeline}\textbf{B}).
	
	\paragraph{Experimental setup.}	We partition the development data using five-fold cross-validation, stratifying jointly by recurrence status, prostatectomy ISUP grade, and whether the radical prostatectomy occurred before or after $2005$, reflecting a change in clinical guidelines that year. We use the open-source Python library \texttt{scikit-multilearn}, which supports stratification across multiple discrete labels. To evaluate the impact of training data diversity on model performance, we maintain two distinct versions of the development data. The first includes $508$ patients from the \texttt{RUMC} cohort, while the second combines these $508$ \texttt{RUMC} patients with the $298$ patients from the \texttt{TCGA-PRAD} cohort. Each version is split independently into five cross-validation folds, referred to as \texttt{RUMC-only} and \texttt{RUMC+TCGA} splits, respectively.

	\section{Human Ethics Approval}
	
	This study was conducted in accordance with the ethical principles of the Declaration of Helsinki and the Netherlands Code of Conduct for Research Integrity.\\
	\\
	The Medical Ethics Committee of Radboud University Medical Center approved the use of the \texttt{RUMC} data in this study on 5/1/2016 (Approval No. 20162275). Contact person: Geert Litjens, Department of Pathology, Radboud Institute for Health Science, Radboud University Medical Center, Nijmegen, The Netherlands (\texttt{geert.litjens@radboudumc.nl}).\\
	\\
	The \texttt{UHC} cohort was created in the course of two studies, each approved by the institutional review board (Approval Nos. 20-1583 and 22-1233). Contact person: Iurii Tolkach, Institute of Pathology, University Hospital Cologne, Cologne, Germany (\texttt{iurii.tolkach@uk-koeln.de}).\\
	\\
	The use of retrospective patient data from \texttt{IMP}, Belo Horizonte, Brazil, was approved by the institutional review board. Contact person: Paulo Guilherme de Oliveira Salles, Anatomical Pathology Service, Instituto Mario Penna, Belo Horizonte, Brazil (\texttt{paulo.salles@mariopenna.org.br}).
	
	\section{Consent to Participate}
	
	For the \texttt{RUMC} cohort, the Medical Ethics Committee of Radboud University Medical Center waived the requirement for informed consent. For the \texttt{UHC} cohort, the data were collected as part of two studies approved by the institutional review board of the University Hospital Cologne. All consent procedures were carried out in accordance with the requirements of the approving committee. The institutional review board approval was waived for all analyses of retrospective, de-identified \texttt{IMP} patient data. Patients were informed at registration and were able to opt out at any time.
	
	\section{Data Availability}
	
	The development cohort from \texttt{RUMC} is publicly available under a CC-BY-NC-SA license via AWS at \href{https://registry.opendata.aws/leopard}{\small\url{https://registry.opendata.aws/leopard}}. The curated dataset of 298 patients from \texttt{TCGA-PRAD} is available under a CC-BY license at \href{https://huggingface.co/datasets/waticlems/tcga-prad-bcr}{\small\url{https://huggingface.co/datasets/waticlems/tcga-prad-bcr}}. Data from \texttt{PLCO} is available upon submission of a project proposal through the Cancer Data Access System (CDAS).
	
	\section{Code Availability}
	
	The code used in this study is publicly available at \href{https://github.com/clemsgrs/bcr-risk-prediction}{\small\url{https://github.com/clemsgrs/bcr-risk-prediction}}. We have also open-sourced our \texttt{Prost40M} model, which can be accessed at \href{https://huggingface.co/waticlems/Prost40M}{\small\url{https://huggingface.co/waticlems/Prost40M}}.
	
	\section*{Acknowledgements} 
	
	The creation of the cohort of University Hospital Cologne was financed by the Wilhelm Sander Foundation, Munich, Germany (Y. Tolkach: grant 2022.040.1) and by Federal Ministry of Education and Research of Germany (BMBF; Project FED-PATH, Y. Tolkach and R. Büttner).\\
	\\
	V. Agosti has been supported by the Beretta Foundation (Italy). E. Munari has been partially supported by the European Union - NextGenerationEU, through the Italian Ministry of University and Research under PNRR - M4C2-I1.3 Project PE\_00000019 "HEAL ITALIA", CUP: B33C22001030006.
	
	\section*{Author Contributions}
	
	C.G. conceptualization, data curation, model development, result analyses; K.F. data curation; N.U. data curation; V.A. data curation; E.M. result analyses; S.F.K.J. data curation; P.G.d.O.S. IMP data curation; Y.T. UHC pathology data preparation; R.B. UHC pathology data preparation; S.S. UHC clinical data preparation; M.P. UHC clinical data preparation; A.H. UHC clinical data preparation. J.v.d.L. conceptualization, supervision; G.L. conceptualization, supervision. All authors reviewed the manuscript. 
	
	\section*{Competing Interests}
	
	J.v.d.L. reports consulting fees from Philips, ContextVision and AbbVie, and grants from Philips, ContextVision and Sectra, outside the submitted work. G.L. reports grants from the Dutch Cancer Society and the NWO, and grants from Philips Digital Pathology Solutions and personal fees from Novartis, outside the submitted work. All other authors declare no relevant competing interests concerning this paper. 
	
	\bibliographystyle{naturemag} 
	\bibliography{references} 
	
	\clearpage
	\appendix

	\section{Statistical Testing}
	\label{app:statistics}
	
	\subsection{Pairwise Model Comparisons}
	\label{app:pairwise-comparisons}
	
	\begin{table}[ht]
		\centering
		\caption{\textbf{Pairwise comparison of encoders trained on \texttt{RUMC-only} data.} The table reports the difference in concordance index ($\Delta$) between pairs of encoders when evaluated on each test cohort. For each comparison, $\Delta$ values are shown together with 95\% bootstrap confidence intervals (CI) and false discovery rate (FDR) adjusted $q$ values. Positive $\Delta$ values indicate higher performance of the encoder listed first in the comparison, whereas negative values indicate higher performance of the encoder listed second. Values that remain significant after correction ($q < 0.05$) are shown in bold.}
		\resizebox{\textwidth}{!}{\begin{tabular}{llccc}
				\toprule
				\textbf{Cohort} & \textbf{Model Comparison} & $\Delta$ & \textbf{95\% CI} & $q$ \\
				\midrule
				\texttt{RUMC} & \texttt{Prost40M} - \texttt{UNI}         & -0.017          & [-0.107, 0.080]      & 0.852 \\
				& \texttt{Prost40M} - \texttt{Virchow2}    & -0.031          & [-0.128, 0.075]      & 0.852 \\
				& \texttt{Prost40M} - \texttt{H-optimus-0} & -0.003          & [-0.103, 0.079]      & 0.977 \\
				& \texttt{UNI} - \texttt{Virchow2}         & -0.014          & [-0.070, 0.042]      & 0.852 \\
				& \texttt{UNI} - \texttt{H-optimus-0}      & +0.014          & [-0.069, 0.086]      & 0.852 \\
				& \texttt{Virchow2} - \texttt{H-optimus-0} & +0.029          & [-0.051, 0.102]      & 0.852 \\
				\midrule
				\texttt{PLCO} & \texttt{Prost40M} - \texttt{UNI}         & \textbf{-0.089} & [-0.138, -0.041]     & $<$ 0.001 \\
				& \texttt{Prost40M} - \texttt{Virchow2}    & \textbf{-0.138} & [-0.181, -0.095]     & $<$ 0.001 \\
				& \texttt{Prost40M} - \texttt{H-optimus-0} & -0.025          & [-0.064, 0.014]      & 0.208 \\
				& \texttt{UNI} - \texttt{Virchow2}         & \textbf{-0.048} & [-0.074, -0.024]     & $<$ 0.001 \\
				& \texttt{UNI} - \texttt{H-optimus-0}      & \textbf{+0.065} & [0.031, 0.101]       & 0.002 \\
				& \texttt{Virchow2} - \texttt{H-optimus-0} & \textbf{+0.113} & [0.083, 0.146]       & $<$ 0.001 \\
				\midrule
				\texttt{IMP}  & \texttt{Prost40M} - \texttt{UNI}         & \textbf{-0.205} & [-0.266, -0.145]     & $<$ 0.001 \\
				& \texttt{Prost40M} - \texttt{Virchow2}    & \textbf{-0.245} & [-0.298, -0.194]     & $<$ 0.001 \\
				& \texttt{Prost40M} - \texttt{H-optimus-0} & \textbf{-0.169} & [-0.215, -0.124]     & $<$ 0.001 \\
				& \texttt{UNI} - \texttt{Virchow2}         & \textbf{-0.040} & [-0.073, -0.008]     & 0.028 \\
				& \texttt{UNI} - \texttt{H-optimus-0}      & +0.036          & [-0.008, 0.083]      & 0.115 \\
				& \texttt{Virchow2} - \texttt{H-optimus-0} & \textbf{+0.076} & [0.044, 0.111]       & $<$ 0.001 \\
				\midrule
				\texttt{UHC}  & \texttt{Prost40M} - \texttt{UNI}         & -0.012          & [-0.054, 0.031]      & 0.608 \\
				& \texttt{Prost40M} - \texttt{Virchow2}    & -0.038          & [-0.079, 0.003]      & 0.148 \\
				& \texttt{Prost40M} - \texttt{H-optimus-0} & +0.015          & [-0.032, 0.063]      & 0.608 \\
				& \texttt{UNI} - \texttt{Virchow2}         & -0.026          & [-0.053, 0.000]      & 0.148 \\
				& \texttt{UNI} - \texttt{H-optimus-0}      & +0.027          & [-0.007, 0.060]      & 0.180 \\
				& \texttt{Virchow2} - \texttt{H-optimus-0} & \textbf{+0.053} & [0.031, 0.075]       & $<$ 0.001 \\
				\bottomrule                                 
			\end{tabular}                                
		}                                             
		\label{table:pairwise-rumc}                   
	\end{table}                               
	
	\begin{table}[ht]
		\centering
		\caption{\textbf{Pairwise comparison of encoders trained on \texttt{RUMC+TCGA} data.} The table reports the difference in concordance index ($\Delta$) between pairs of encoders when evaluated on each test cohort. For each comparison, $\Delta$ values are shown together with 95\% bootstrap confidence intervals (CI) and false discovery rate (FDR) adjusted $q$ values. Positive $\Delta$ values indicate higher performance of the encoder listed first in the comparison, whereas negative values indicate higher performance of the encoder listed second. Values that remain significant after correction ($q < 0.05$) are shown in bold.}
		\resizebox{\textwidth}{!}{\begin{tabular}{llccc}
				\toprule
				\textbf{Cohort} & \textbf{Model Comparison} & $\Delta$ & \textbf{95\% CI} & $q$ \\
				\midrule
				\texttt{RUMC} & \texttt{Prost40M} - \texttt{UNI}         & +0.051          & [-0.021, 0.141]      & 0.283 \\
				& \texttt{Prost40M} - \texttt{Virchow2}    & -0.007          & [-0.124, 0.115]      & 0.882 \\
				& \texttt{Prost40M} - \texttt{H-optimus-0} & +0.088          & [0.005, 0.183]       & 0.151 \\
				& \texttt{UNI} - \texttt{Virchow2}         & -0.058          & [-0.153, 0.024]      & 0.283 \\
				& \texttt{UNI} - \texttt{H-optimus-0}      & +0.037          & [-0.056, 0.137]      & 0.548 \\
				& \texttt{Virchow2} - \texttt{H-optimus-0} & +0.096          & [-0.000, 0.201]      & 0.151 \\
				\midrule
				\texttt{PLCO} & \texttt{Prost40M} - \texttt{UNI}         & +0.037          & [0.000, 0.077]       & 0.114 \\
				& \texttt{Prost40M} - \texttt{Virchow2}    & +0.036          & [-0.001, 0.076]      & 0.114 \\
				& \texttt{Prost40M} - \texttt{H-optimus-0} & \textbf{+0.060} & [0.022, 0.102]       & 0.024 \\
				& \texttt{UNI} - \texttt{Virchow2}         & -0.002          & [-0.035, 0.033]      & 0.943 \\
				& \texttt{UNI} - \texttt{H-optimus-0}      & +0.023          & [-0.009, 0.057]      & 0.198 \\
				& \texttt{Virchow2} - \texttt{H-optimus-0} & +0.024          & [-0.009, 0.059]      & 0.198 \\
				\midrule
				\texttt{IMP}  & \texttt{Prost40M} - \texttt{UNI}         & +0.031          & [-0.011, 0.075]      & 0.216 \\
				& \texttt{Prost40M} - \texttt{Virchow2}    & -0.007          & [-0.060, 0.048]      & 0.793 \\
				& \texttt{Prost40M} - \texttt{H-optimus-0} & \textbf{+0.137} & [0.079, 0.197]       & $<$ 0.001 \\
				& \texttt{UNI} - \texttt{Virchow2}         & -0.038          & [-0.097, 0.023]      & 0.271 \\
				& \texttt{UNI} - \texttt{H-optimus-0}      & \textbf{+0.106} & [0.038, 0.174]       & 0.003 \\
				& \texttt{Virchow2} - \texttt{H-optimus-0} & \textbf{+0.144} & [0.092, 0.196]       & $<$ 0.001 \\
				\midrule
				\texttt{UHC}  & \texttt{Prost40M} - \texttt{UNI}         & +0.006          & [-0.028, 0.042]      & 0.901 \\
				& \texttt{Prost40M} - \texttt{Virchow2}    & -0.041          & [-0.078, -0.003]     & 0.070 \\
				& \texttt{Prost40M} - \texttt{H-optimus-0} & +0.008          & [-0.031, 0.046]      & 0.901 \\
				& \texttt{UNI} - \texttt{Virchow2}         & \textbf{-0.047} & [-0.079, -0.018]     & 0.009 \\
				& \texttt{UNI} - \texttt{H-optimus-0}      & +0.002          & [-0.030, 0.034]      & 0.903 \\
				& \texttt{Virchow2} - \texttt{H-optimus-0} & \textbf{+0.049} & [0.015, 0.084]       & 0.011 \\
				\bottomrule                                 
			\end{tabular}                                
		}                                             
		\label{table:pairwise-rumc+tcga}                   
	\end{table}
	
	\clearpage
	
	\subsection{Effect of Training Dataset Enrichment}
	\label{app:dataset-enrichement}
	
	\begin{table}[ht]
		\centering
		\caption{\textbf{Effect of training data enrichment on model performance.} The table shows the change in concordance index ($\Delta$) between models trained on the combined \texttt{RUMC+TCGA} splits and those trained on \texttt{RUMC-only} splits. For each test cohort and encoder, $\Delta$ values are reported alongside 95\% bootstrap CI, the exact two-sided $p$ values, and FDR-adjusted $q$ values. Positive $\Delta$ values indicate higher performance after training data enrichment, whereas negative values indicate lower performance. Values that remain significant after FDR correction ($q < 0.05$) are highlighted in bold.}
		\begin{tabular}{llcccc}
			\toprule
			\textbf{Cohort} & \textbf{Encoder} & $\Delta$ & \textbf{95\% CI }& $p$ & $q$ \\
			\midrule
			\texttt{RUMC} & \texttt{Prost40M}    & \textbf{+0.074} & [0.023, 0.136] & 0.002      & 0.008 \\
			& \texttt{UNI}         & +0.006          & [-0.060, 0.067] & 0.866      & 0.990 \\
			& \texttt{Virchow2}    & +0.050          & [-0.019, 0.139] & 0.190      & 0.380 \\
			& \texttt{H-optimus-0} & -0.017          & [-0.112, 0.075] & 0.708      & 0.871 \\
			\midrule
			\texttt{PLCO} & \texttt{Prost40M}    & \textbf{+0.107} & [0.052, 0.161] & $<$ 0.001  & $<$ 0.001 \\
			& \texttt{UNI}         & -0.020          & [-0.064, 0.021] & 0.353      & 0.565 \\
			& \texttt{Virchow2}    & \textbf{-0.067} & [-0.103, -0.033] & $<$ 0.001  & $<$ 0.001 \\
			& \texttt{H-optimus-0} & +0.022          & [-0.031, 0.072] & 0.414      & 0.602 \\
			\midrule
			\texttt{IMP}  & \texttt{Prost40M}    & \textbf{+0.238} & [0.159, 0.318] & $<$ 0.001  & $<$ 0.001 \\
			& \texttt{UNI}         & +0.002          & [-0.050, 0.053] & 0.933      & 0.993 \\
			& \texttt{Virchow2}    & 0.000           & [-0.032, 0.031] & 0.993      & 0.993 \\
			& \texttt{H-optimus-0} & -0.068          & [-0.130, -0.006] & 0.033      & 0.104 \\
			\midrule
			\texttt{UHC}  & \texttt{Prost40M}    & +0.031          & [-0.024, 0.083] & 0.260      & 0.462 \\
			& \texttt{UNI}         & +0.014          & [-0.040, 0.064] & 0.593      & 0.790 \\
			& \texttt{Virchow2}    & +0.034          & [-0.000, 0.068] & 0.052      & 0.139 \\
			& \texttt{H-optimus-0} & +0.038          & [-0.012, 0.084] & 0.133      & 0.303 \\
			\bottomrule
		\end{tabular}
		\label{table:dataset-enrichement-statistics}
	\end{table}
	
	\clearpage
	
	\section{Ensemble and joint modeling performance of models trained on \texttt{RUMC-only} splits}
	\label{app:joint-modeling-rumc}
	
	\begin{table}[htbp]
		\centering
		\caption{\textbf{Ensemble and joint modeling performance of models trained on \texttt{RUMC-only}.} Summary of the prognostic performance of deep learning models evaluated on the four held-out test cohorts. For each cohort, the \emph{ensemble} c-index reflects the standalone performance of the deep learning model and is computed by averaging patient-level risk scores from the five \texttt{RUMC-only} cross-validation folds. The \emph{joint} c-index is obtained by fitting a multivariable CoxPH model using both the ensemble risk score and the CAPRA-S score as covariates. For joint models, the hazard ratio (HR) with $95$\% confidence interval (CI) and the Wald test $p$-value indicate the independent contribution of the ensemble risk score after adjusting for CAPRA-S. Bold values indicate the highest c-index observed for each cohort.}
		\resizebox{\textwidth}{!}{
			\begin{tabular}{llccccc}
				\toprule
				\textbf{Test set} & \textbf{Encoder} & \textbf{Ensemble} &  \textbf{Joint} & \textbf{HR (95\% CI)} & \textbf{p-value} \\
				\midrule
				\multirow{4}{*}{\texttt{RUMC}} & \texttt{Prost40M}    & 0.686           & 0.754           & 1.50 (1.05-2.15)     & 0.028 \\
				& \texttt{UNI}         & 0.703           & 0.739           & 1.31 (0.93-1.85)     & 0.122 \\
				& \texttt{Virchow2}    & \textbf{0.717}  & \textbf{0.759}  & 1.50 (1.04-2.17)     & 0.030 \\
				& \texttt{H-optimus-0} & 0.688           & 0.733           & 1.17 (0.86-1.60)     & 0.327 \\
				\midrule
				\multirow{4}{*}{\texttt{PLCO}} & \texttt{Prost40M}    & 0.604           & 0.766           & 1.07 (0.98-1.17)     & 0.121 \\
				& \texttt{UNI}         & 0.694           & 0.769           & 1.26 (1.11-1.43)     & $<$ 0.001 \\
				& \texttt{Virchow2}    & \textbf{0.742}  & \textbf{0.778}  & 1.27 (1.13-1.43)     & $<$ 0.001 \\
				& \texttt{H-optimus-0} & 0.629           & 0.763           & 1.18 (1.04-1.32)     & 0.007 \\
				\midrule
				\multirow{4}{*}{\texttt{IMP}}  & \texttt{Prost40M}    & 0.462           & 0.771           & 0.88 (0.75-1.03)     & 0.123 \\
				& \texttt{UNI}         & 0.667           & 0.775           & 1.05 (0.92-1.20)     & 0.497 \\
				& \texttt{Virchow2}    & \textbf{0.707}  & \textbf{0.779}  & 1.17 (1.01-1.36)     & 0.037 \\
				& \texttt{H-optimus-0} & 0.631           & 0.774           & 1.10 (0.93-1.29)     & 0.271 \\
				\midrule
				\multirow{4}{*}{\texttt{UHC}}  & \texttt{Prost40M}    & 0.663           & \textbf{0.750}  & 1.04 (0.93-1.16)     & 0.528 \\
				& \texttt{UNI}         & 0.675           & 0.748           & 1.20 (1.00-1.44)     & 0.052 \\
				& \texttt{Virchow2}    & \textbf{0.701}  & 0.747           & 1.26 (1.04-1.54)     & 0.018 \\
				& \texttt{H-optimus-0} & 0.648           & 0.746           & 1.08 (0.90-1.29)     & 0.412 \\
				\bottomrule
			\end{tabular}
			\label{table:multivariable-cox-models-rumc}
		}
	\end{table}
	
	\clearpage
	
	\section{ISUP-stratified performance on UHC}
	\label{app:subgroup-analysis-uhc}
	
	The \texttt{UHC} cohort consists of $330$ cases, distributed as follows: 22 ISUP 1, 127 ISUP 2, 86 ISUP 3, 21 ISUP 4, and 74 ISUP 5.
	
	\begin{figure}[!htbp]
		\centering
		\includegraphics[width=\linewidth]{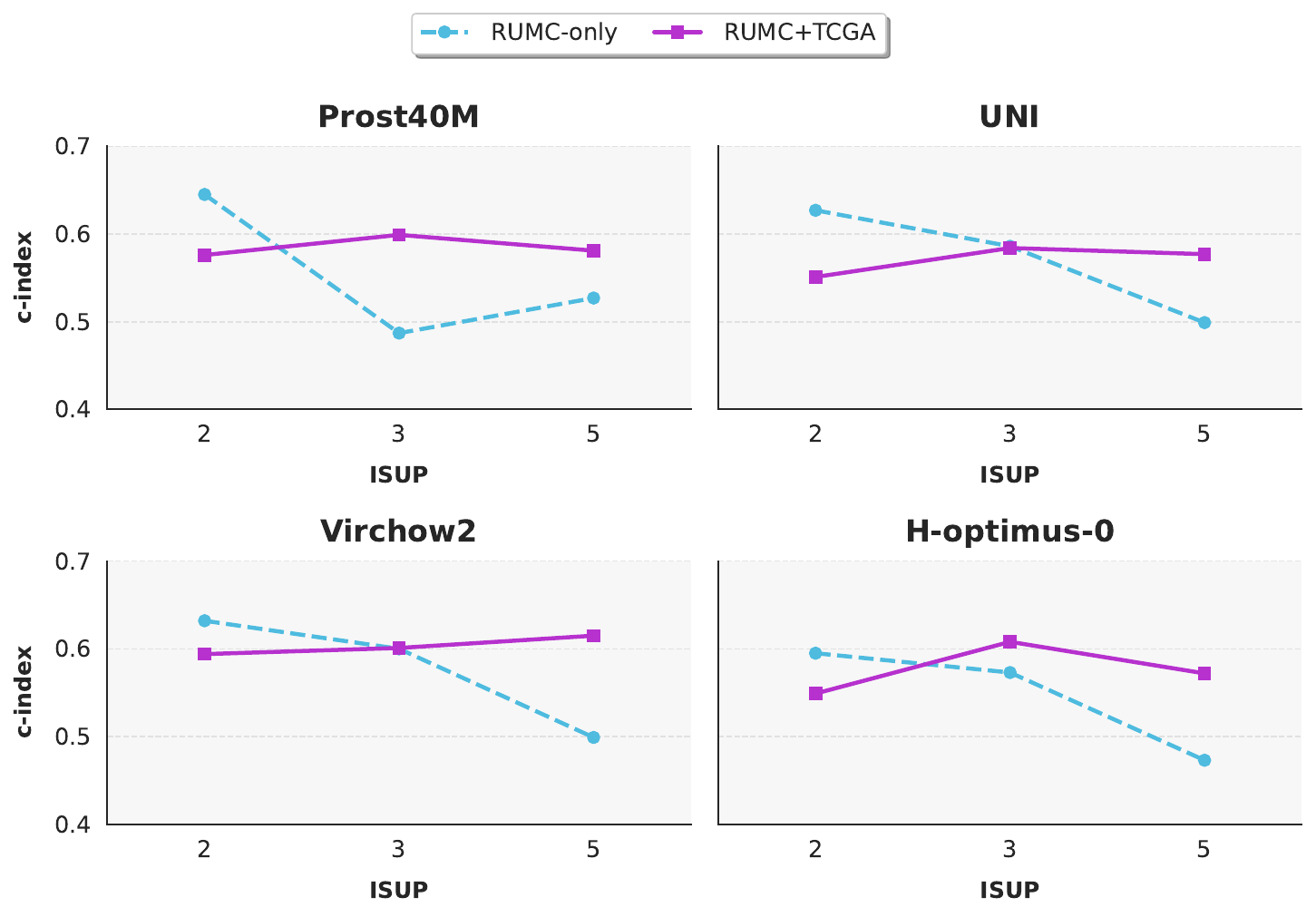}
		\caption{\textbf{ISUP-stratified predictive performance on the \texttt{UHC} cohort before and after training set enrichment.} Ensemble c-index of models trained on \texttt{RUMC-only} versus \texttt{RUMC+TCGA}, stratified by ISUP grade within the \texttt{UHC} cohort. Results exclude grade 1 (all censored, preventing c-index calculation) and grade 4 (small sample size). Models trained on \texttt{RUMC-only} show performance skewed toward ISUP 2, reflecting the grade distribution in the training set. Adding \texttt{TCGA-PRAD} cases reduces this bias, improving performance in higher-grade cases (ISUP 3 and 5) and yielding a flatter profile across grades, consistent with better calibration of recurrence risk across grades.}
		\label{fig:subgroup-analysis-uhc}
	\end{figure}
	
	\clearpage
	
	\section{Contribution cluster analysis}
	\label{app:cluster-analysis}
	
	For three risk-increasing and two risk-lowering clusters, we present grids of representative tiles. Each grid displays individual tiles highlighted within their surrounding whole-slide context, facilitating interpretation of the histomorphological patterns that the model associates with increased or reduced recurrence risk.
	
	\subsection{Risk-increasing clusters}
	
	\begin{figure}[!htbp]
		\centering
		\includegraphics[width=\textwidth]{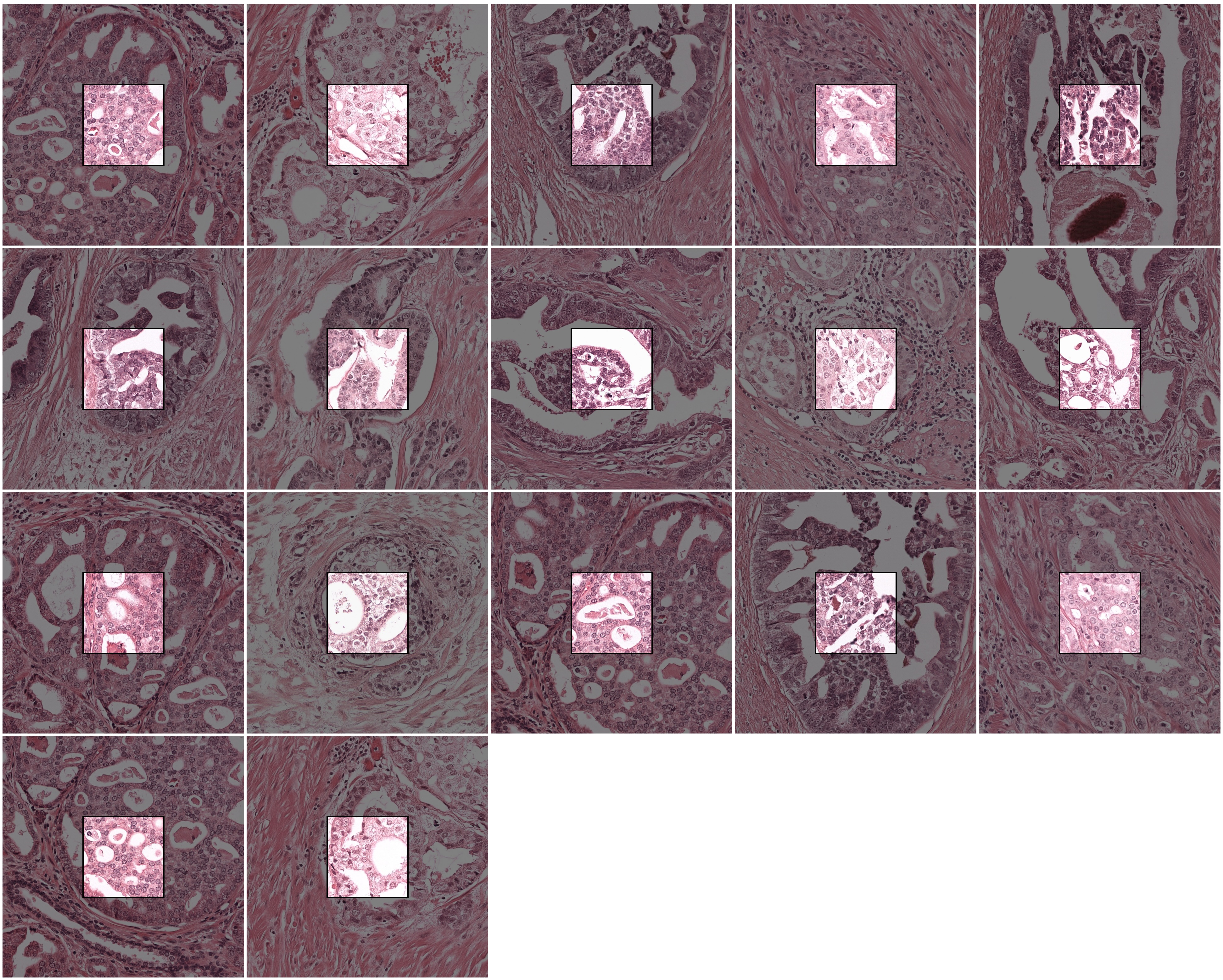}
		\caption{\textbf{Representative tiles from risk-increasing cluster 1.} This cluster is dominated by adverse histology, including extensive cribriform structures and fused glands forming anastomosing cords, aligning with the model’s interpretation of this cluster as increasing risk of recurrence.}
		\label{fig:tsne_high_cluster1}
	\end{figure}
	
	\begin{figure}[!htbp]
		\centering
		\includegraphics[width=\textwidth]{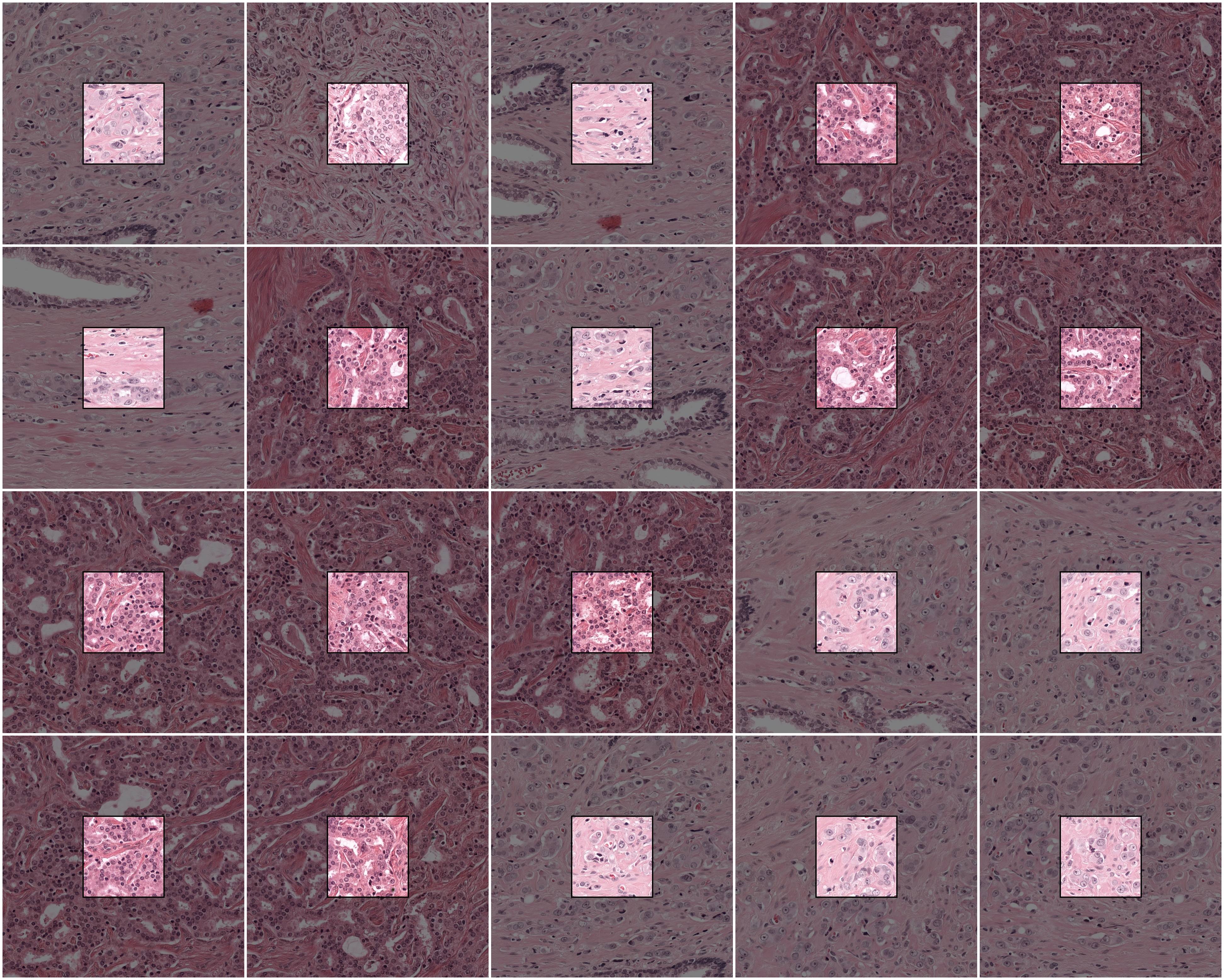}
		\caption{\textbf{Representative tiles from risk-increasing cluster 2.} Tiles predominantly contain high-grade carcinoma, including single infiltrating cells characteristic of Gleason pattern 5 and fused pattern 4 glands forming anastomosing cords. These morphologies are strongly prognostically adverse and align with large positive contributions to risk of recurrence.}
		\label{fig:tsne_high_cluster2}
	\end{figure}
	
	\begin{figure}[!htbp]
		\centering
		\includegraphics[width=\textwidth]{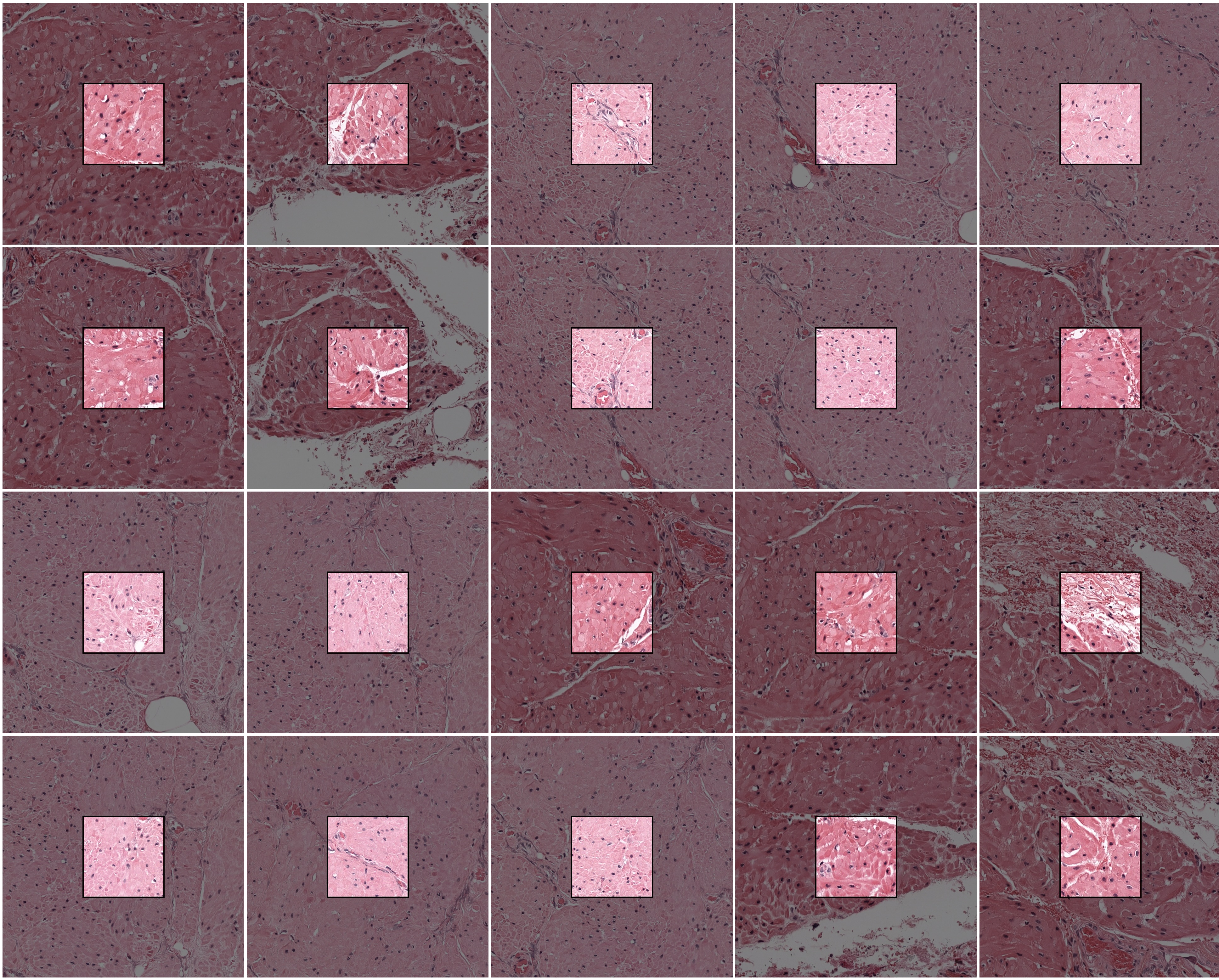}
		\caption{\textbf{Representative tiles from risk-increasing cluster 3.} Tiles primarily depict fibromuscular stroma rather than carcinoma, suggesting that the model may partially attribute recurrence risk based on patient- or slide-specific non-tumor features.}
		\label{fig:tsne_high_cluster3}
	\end{figure}
	
	\clearpage
	
	\subsection{Risk-lowering clusters}
	
	\begin{figure}[!htbp]
		\centering
		\includegraphics[width=\textwidth]{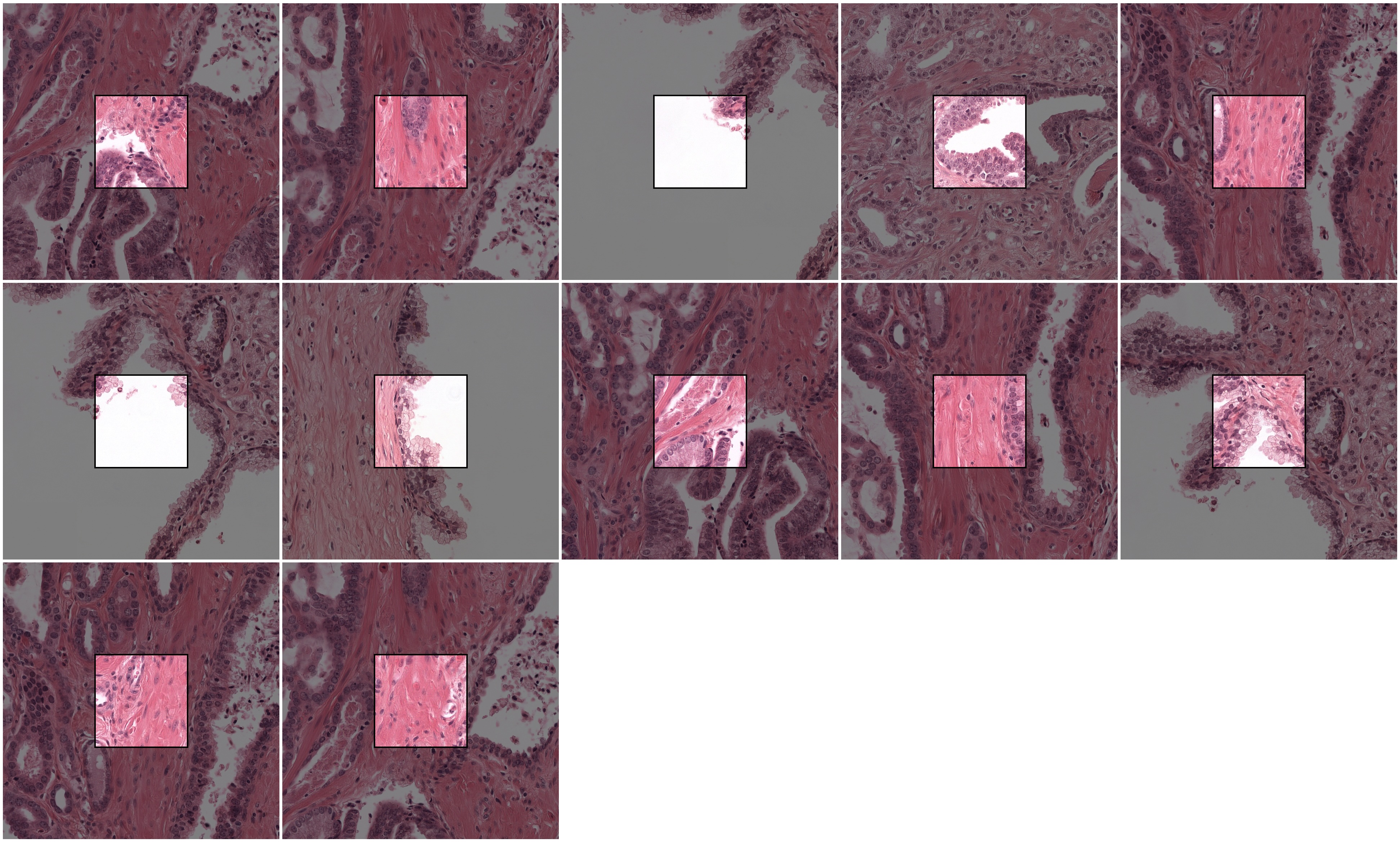}
		\caption{\textbf{Representative tiles from risk-lowering cluster 1.} Most tiles display predominantly benign tissue with only occasional proximity to carcinoma.}
		\label{fig:tsne_low_cluster1}
	\end{figure}
	
	\begin{figure}[!htbp]
		\centering
		\includegraphics[width=\textwidth]{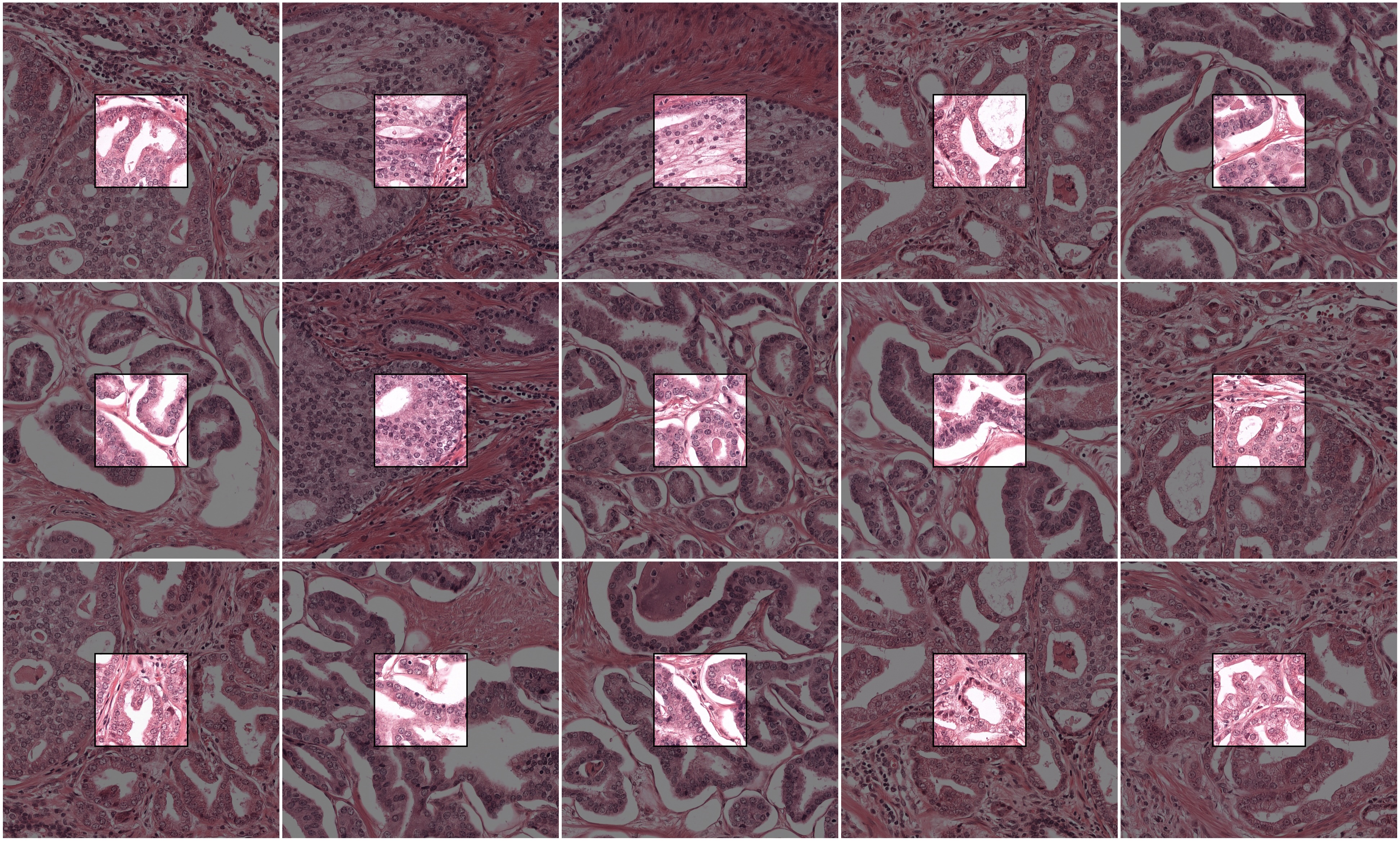}
		\caption{\textbf{Representative tiles from risk-lowering cluster 2.} While some tiles show cribriform morphology which should confer higher risk, the overall Gleason pattern 4 complexity seen in other tiles is more consistent with a favorable histological appearance.}
		\label{fig:tsne_low_cluster2}
	\end{figure}
	
	\clearpage
	
	\section{Dataset details}
	\label{app:datasets}
	
	We provide detailed inclusion and exclusion criteria, scanner specifications, tumor subtype considerations, BCR label definitions, and follow-up data extraction procedures for all datasets used in this study.
	
	\subsection{RUMC}
	
	The Radboud University Medical Center (RUMC, Nijmegen, The Netherlands) cohort initially consisted of $903$ patients who underwent radical prostatectomy between $1992$ and $2012$. Patients were monitored for biochemical recurrence through PSA follow-up, typically within one year at the hospital and thereafter by general practitioners. The following cases were excluded: no consent ($2$), no follow-up information ($22$), PSA levels failing to reach nadir after prostatectomy ($83$), history of neoadjuvant treatment ($12$), absence of sufficiently digitizable slides ($120$), unclear tumor location ($6$), and predominant ductal carcinoma subtype ($1$). This resulted in a final cohort of $657$ patients. Of these, $49$ were used as an intermediate evaluation cohort during the \texttt{LEOPARD} challenge and are excluded from this study. We used the remaining $608$ patients: $508$ publicly released development cases and $100$ held-out test cases.
	
	\subsection{TCGA-PRAD}
	
	TCGA includes multiple recurrence-related fields that were found to be inconsistent across data releases. To construct reliable BCR labels, we curated recurrence status and event times using case-specific clinical XML files. A recurrence event was assigned if there was either $(1)$ a positive BCR indicator with a corresponding date of recurrence, $(2)$ evidence of biochemical disease progression recorded as a new tumor event, or $(3)$ a recurrence date provided without explicit indicator status. Patients without any sign of recurrence were labeled as censored. Patients with ambiguous or conflicting information were excluded. Through this process, we identified $298$ patients with BCR labels, including $62$ recurrence events and $236$ censored cases. These patients were linked to $332$ WSIs. Although the remaining $105$ patients had incomplete or unreliable recurrence information, their slides were retained for self-supervised pretraining. In total, this resulted in $403$ patients and $449$ slides from \texttt{TCGA-PRAD} used in this study.
	
	\subsection{PLCO}
	
	The Prostate, Lung, Colorectal, and Ovarian (PLCO) Cancer Screening Trial enrolled approximately $155,000$ participants between $1993$ and $2001$, with follow-up through December $31$, $2009$. We identified patients with BCR using the \texttt{rp\_bcr\_0p2consec} variable, for which we could get time to BCR as the time from surgery to two consecutive positive PSA measurements (\texttt{rp\_bcr\_0p2consec\_mo}). For patients without BCR, time to censoring or last follow-up was measured from the date of surgery (calculated as \texttt{primary\_trtp\_days} - \texttt{candxdaysp}) to the date of study exit (\texttt{pcp\_bcr\_exit\_days}). Cases with reported distant metastasis present at the time of surgery $(1)$ and negative time to follow-up $(2)$ were excluded. In the end, we obtained $773$ patients with at least one digital pathology slide. All slides had been scanned using a Leica scanner at a resolution of $0.25$ mpp. Like the \texttt{RUMC} cohort, these cases were included in the \texttt{LEOPARD} challenge, where $50$ were used for intermediate evaluation and $723$ were held out for final testing. In this study, we only used the $723$ hold-out cases for testing, excluding the $50$ cases used during the challenge evaluation phase.
	
	\subsection{IMP}
	
	The Instituto Mário Penna (IMP) cohort initially consisted of $437$ patients who underwent radical prostatectomy in $2016$. Patients were all followed up until $2022$: every three months during the first year, every four months in the second year, every six months during the third and fourth years, and annually from the fifth year onward. No patients had documented distant metastasis at the time of surgery or received neoadjuvant hormonal therapy. In all cases, the predominant cancer component was not ductal adenocarcinoma. We excluded cases without any digitized slide $(10)$, and cases where PSA levels failed to reach a nadir after radical prostatectomy $(6)$, resulting in a final cohort of $421$ patients. Histopathology slides for each case were digitized using a Motic Easy Scan Infinity $60$N scanner at a resolution of $0.26$ mpp, resulting in a total of $2091$ digital prostatectomy slides.
	
	\subsection{UHC}
	
	The University Hospital Cologne (UHC) cohort initially consisted of $337$ patients who underwent radical prostatectomy between $2015$ and $2020$ in Cologne, Germany. All patients had no history of hormonal therapy or evidence of therapy-induced regression, as per the cohort’s predefined inclusion criteria. We further excluded cases where the majority of the tumor was classified as ductal adenocarcinoma $(3)$, cases with negative or missing time to recurrence $(3)$, and cases where PSA levels failed to reach a nadir after radical prostatectomy $(1)$, resulting in a final cohort of $330$ patients. Histopathology slides for each case were digitized using a Hamamatsu NanoZoomer S$360$ scanner at a resolution of $0.23$ mpp, resulting in a total of $1028$ digital prostatectomy slides.\\
	\\
	Importantly, the definition of BCR differed in the \texttt{UHC} cohort compared with the other datasets. Whereas \texttt{RUMC}, \texttt{PLCO}, and \texttt{IMP} adopt the historical standard of two consecutive PSA measurements $\geq 0.2$ ng/mL, \texttt{UHC} defines BCR as a single postoperative PSA $\geq 0.1$ ng/mL. This shift reflects the evolution of clinical practice and improvements in PSA assay sensitivity, with the second confirmatory test rarely altering classification. Another key difference is the distribution of tumor grades: the \texttt{UHC} cohort contains relatively few low-grade (ISUP 1) prostate cancers. This again reflects changes in treatment guidelines: active surveillance is now commonly recommended for low-grade disease, resulting in fewer such patients undergoing surgery. The grade distribution in \texttt{UHC} is more similar to that of the \texttt{TCGA-PRAD} cohort, and contrasts with the \texttt{RUMC}, \texttt{PLCO}, and \texttt{IMP} cohorts, which include a higher proportion of low-grade cases (Figure \ref{fig:isup}). While introducing heterogeneity, these differences make the \texttt{UHC} cohort particularly valuable for testing model generalization to contemporary clinical practice.
	
\end{document}